\documentclass{article}
\usepackage{ijcai24}

\usepackage{times}
\usepackage{soul}
\usepackage{url}
\usepackage[hidelinks]{hyperref}
\usepackage[utf8]{inputenc}
\usepackage[small]{caption}
\usepackage{graphicx}
\usepackage{amsmath}
\usepackage{amsthm}
\usepackage{booktabs}
\usepackage{amssymb}
\usepackage{algorithm}
\usepackage{algorithmic}
\usepackage[switch]{lineno}
\usepackage{multirow}
\usepackage{color}
\usepackage{bbding}
\usepackage{float}
\usepackage{alltt}
\usepackage{diagbox}
\usepackage{hyperref}
\usepackage[dvipsnames]{xcolor,colortbl}

\newtheorem{proposition}{Proposition}

\title{Dynamic Temperature Knowledge Distillation}

\author{
   Yukang Wei, Yu Bai
   \affiliations
   HEBUSTS
   \emails
   weiyukang1998.gm@gmail.com,
   baiyu@hebust.edu.cn
}

\begin{document}

\maketitle

\begin{abstract}

Temperature plays a pivotal role in moderating label softness in the realm of knowledge distillation (KD). Traditional approaches often employ a static temperature throughout the KD process, which fails to address the nuanced complexities of samples with varying levels of difficulty and overlooks the distinct capabilities of different teacher-student pairings. This leads to a less-than-ideal transfer of knowledge. To improve the process of knowledge propagation, we proposed Dynamic Temperature Knowledge Distillation (DTKD) which introduces a dynamic, cooperative temperature control for both teacher and student models simultaneously within each training iterafion.  In particular, we proposed ``\textbf{sharpness}'' as a metric to quantify the smoothness of a model's output distribution. By minimizing the sharpness difference between the teacher and the student, we can derive sample-specific temperatures for them respectively. Extensive experiments on CIFAR-100 and ImageNet-2012 demonstrate that DTKD performs comparably to leading KD techniques, with added robustness in Target Class KD and None-target Class KD scenarios.The code is available at \href{https://github.com/JinYu1998/DTKD}{this https URL}.

\end{abstract}

\setlength{\abovedisplayskip}{3pt}
\setlength{\belowdisplayskip}{3pt}

\section{Introduction}
Knowledge distillation \cite{hinton2015distilling} is a very effective model compression technology that aims to transfer knowledge from a more bloated and complex teacher model to a more compact and lightweight student model so that the model can be deployed on devices with more restricted resources. Typically the output of the teacher is too sharp (confident) and makes it difficult for the student model to learn the subtle differences between the incorrect classes. As a result, soft labels regulated by ``temperature'' are often used to improve the KD performance. In practice, the loss function of knowledge distillation usually consists of two parts: cross-entropy (CE) loss with hard labels, and KL-divergence loss with soft labels. Temperature has a great impact on the smoothness of the soft labels. Larger temperatures make the soft label smoother, and smaller temperatures make the soft label sharper.

\begin{figure}[H]
    \centering
    \includegraphics[width=\linewidth]{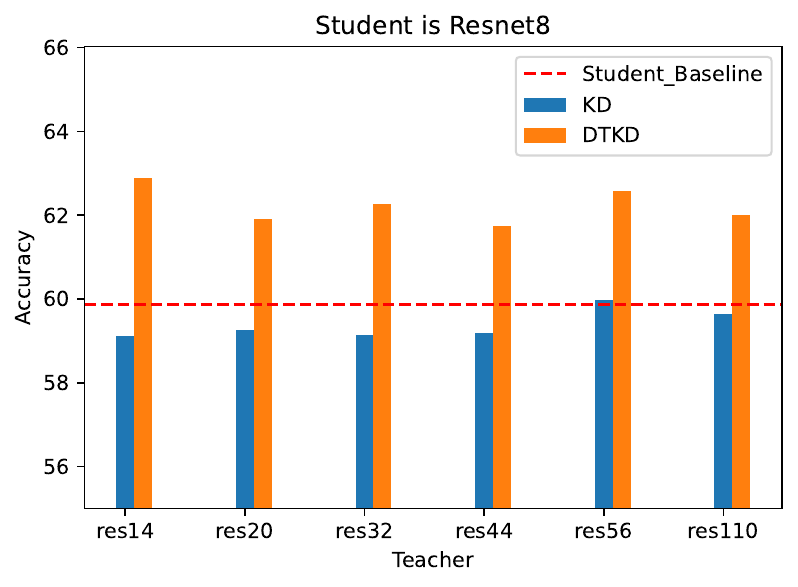}
    \caption{ResNet8 as student tries to learn from teacher models of different sizes. The figure shows the student accuracy of the base line model, the vanilla KD and our DTKD. Both the fixed temperature of KD and the reference DTKD temperature are $4.0$.}
    \label{fig:student_resnet8}
\end{figure}

A few research studies have observed that a fixed temperature might hinder the KD process, and therefore devoted to exploring the dynamic regulation of temperature during the KD process. For instance, Annealing KD \cite{jafari2021annealing} believed that tuning the temperature can bridge the capacity gap between the teacher and the student networks, and proposed to control the temperature through simulated annealing. 
CTKD \cite{li2023curriculum} showed that the task difficulty level can be controlled during the student’s learning career through a dynamic and learnable temperature.
These methods overlooked the differences in smoothness between the output distributions of the student and teacher networks, and they all apply the same temperature to both the teacher and the student. 
Furthermore, all the above methods need to add new modules or increase training costs, which greatly reduces the flexibility of their usage. As a result, empirically fixed temperatures remain in most recently developed KD algorithms (for example, DKD\cite{zhao2022decoupled}). NormKD \cite{chi2023normkd} originates from a perspective akin to ours, employing normalization of the output logits to establish a consistent degree of smoothness across the distribution. However, their approach does not account for the collaborative dynamics between teacher and student models, resulting in students with markedly low confidence levels that render them less practical for real-world deployment.

Different from most previous works, our work is based on the following observation: if a single fixed temperature is shared between the teacher and the student, there usually exists a difference in the smoothness of their logits (output of the network) which could hinder the KD process. This phenomenon becomes more pronounced when the student's capability is limited. As shown in \autoref{fig:student_resnet8}, a ResNet8 student is paired with teachers of various sizes, the effect of vanilla KD at temperature $4.0$ is generally worse than the baseline. 
To tackle this mismatch of smoothness, we introduced Dynamic Temperature Knowledge Distillation (DTKD) in this paper. In particular, we proposed to use the $\mathit{logsumexp}$ function to quantify the \textbf{sharpness} of the logits distribution. The difference between sharpness therefore reflects the difference between the smoothness of logits distributions. 
By minimizing the sharpness difference, we can derive sample-wise temperatures for the teacher and the student respectively and collaboratively during each training iteration. As a result, a moderate common ground of softness between the teacher and the student is settled, which helps the student learn from the teacher more effectively. Intensive experiments show that DTKD effectively improves KD performance. In particular, with DTKD at the same reference temperature as KD, the performance of the same ResNet8 student is well above the baseline, as shown in Figure~\ref{fig:student_resnet8}. This indicates that it is indeed the mismatch of smoothness rather than the student's limited learning capability that hinders the distillation.

In summary, our contributions are as follows:
\begin{itemize}
    \item {We investigated the influence of the difference in smoothness of logits of teacher and student networks on KD. Such difference generally exists in the logits regarding each sample, and hinders the KD process.}
    \item We proposed $\mathit{logsumexp}$ function to quantify the \textit{sharpness} of logits, and introduced the Dynamic Temperature Knowledge Distillation {based on minimization of sharpness difference} to obtain the temperatures of teachers and students respectively for each sample in each iteration.
    \item Compared with previous temperature regulation methods, our method finds suitable temperatures for both the teacher and the student collaboratively with ignorable added calculation and can mitigate problems of both model gaps and task difficulties simultaneously.
    \item Extensive experimental results show that our algorithm can be well combined with KD and DKD\cite{zhao2022decoupled} and reaches SOTA in performance. {When DTKD only distills the Target Class or the None-Target Class, the results show more robustness than DKD.} 
\end{itemize}

\section{Related Work}

Knowledge Distillation (KD), a concept introduced by Hinton et al., has significantly impacted deep learning. It involves transferring knowledge from a large `teacher' model to a smaller `student' model. The approaches within KD are diverse, with logit distillation \cite{hinton2015distilling}, \cite{li2023curriculum}, \cite{zhao2022decoupled}  and feature distillation \cite{romero2014fitnets}, \cite{chen2021distilling}, \cite{tian2019contrastive} being the primary focuses. Logit distillation is popular due to its straightforward application, focusing on the final output logits of the models. Our work also follows this approach. The major focus in logit distillation has been to alleviate the capability gap between teachers and students. For instance, DML \cite{zhang2018deep} proposed a novel mutual learning strategy where both teacher and student models learn concurrently. ATKD's \cite{mirzadeh2020improved} introduction of a `teaching assistant' model helps mitigate the complexity gap between the teacher and student models. SimKD's \cite{chen2022knowledge} strategy of integrating the teacher's classifier into the student model, although effective, deviates from the norm by altering the student model's structure. Generally, the performance of logit distillation fell short compared to feature-based methods. However, recent works like DKD \cite{zhao2022decoupled} showed more potential in this approach.

\begin{table*}[t]
    \centering
    \begin{tabular}{c|cc|c|c|c|c}
        \toprule
        Method      &Extra Training &Extra Module& Consistent smoothness & Synergy & Model Variety & Task Difficulty\\
        \midrule
        Annealing KD & \Checkmark &            & \XSolidBrush & \XSolidBrush  & \Checkmark    & \Checkmark \\
        MKD          & \Checkmark & \Checkmark & \XSolidBrush & \Checkmark  & \Checkmark    & \Checkmark \\
        CTKD         &            & \Checkmark & \XSolidBrush & \XSolidBrush  & \Checkmark  & \Checkmark \\
        NormKD       &            &            & \Checkmark   & \XSolidBrush  & \XSolidBrush  & \XSolidBrush \\
        DTKD         &            &            & \Checkmark   & \Checkmark    & \Checkmark    & \Checkmark \\
        \bottomrule
    \end{tabular}
    \caption{Summary of important features of related works in dynamic temperature regulation.}
    \label{tab:compare}
\end{table*}

Recent advancements also introduced sophisticated methods to enhance the knowledge transfer process\cite{li2022asymmetric}, \cite{tang2020understanding}, \cite{yuan2020revisiting}, \cite{zhou2021rethinking}, \cite{li2017learning}. Among these works, a few look for ways to customize temperatures dynamically. Annealing KD \cite{jafari2021annealing} proposes a dynamic temperature term to bridge the gap between teacher and student networks, focusing on an annealing function applied to the teacher's output in KD loss. CTKD\cite{li2023curriculum}adopts a curriculum learning strategy, where the temperature is dynamically adjusted during the training process to organize the distillation from easy to hard. Meta KD (MKD)\cite{liu2022meta} employs meta-learning to automatically adapt temperatures to various datasets and teacher-student configurations. Normalized KD (NormKD)\cite{chi2023normkd} innovates by customizing the temperature for each sample based on its logit distribution, ensuring a more uniform softening of logit outputs. 

In Table~\ref{tab:compare}, we compared the previous temperature regulation methods with our method in five ways: (1) Whether additional resources are needed to assist the distillation, (2) Consistent smoothness, whether the method unifies smoothness for both peers, (3) Temperature synergy, which refers to whether the method takes into account the cooperation between the teacher and the student, (4) Model variety: whether the gap between various pairs of student and teacher are taken into account, and (5) Task difficulty, whether the method takes into account the task difficulty as training goes on.

\section{Methodology}

\subsection{Background}
Vanilla KD only concentrates on aligning the logit outputs between the teacher and the student. For a given training sample $\mathbf x$ accompanied by a one-hot ground truth label $\mathbf y$, the corresponding output logits  $\mathbf u$ and $\mathbf v$ are obtained from the teacher and student networks respectively. The vector $\mathbf{u} = [u_1, u_2, ..., u_K] \in \mathbb{R}^{1\times K}$ consists of elements $u_i$, each representing the teacher network's output logit for the i-th class, where $K$ denotes the total number of classes in the dataset. It is trivial that $\mathbf v$ follows the same structure. The classification probabilities corresponding to $\mathbf u$ and $\mathbf v$ after $\textit{softmax}$ are $\mathbf{p} = [p_1, p_2, ...,p_K]$ and $\mathbf{q} = [q_1, q_2, ..., q_K]$ respectively. In vanilla KD, the  $p_i$ and $q_i$ are softened by a fixed temperature $\tau$ during the $\mathit{softmax}$ computation as follows:

\vspace{-4mm}
\begin{equation}
\begin{aligned}
    p_{i} = \frac{\exp(u_i / \tau)}{\sum_{i=1}^K \exp(u_i / \tau)}   \quad\quad\quad
    q_{i} = \frac{\exp(v_i / \tau)}{\sum_{i=1}^K \exp(v_i / \tau)} 
\end{aligned}
\end{equation}

Apparently, the temperature $\tau$ controls the degree of softening of the probability. When $\tau$ is $1.0$, the above calculations turn to the vanilla \textit{softmax} operation. The final KD loss function consists of a cross-entropy loss ($\tau=1.0$) and a KL-Divergence loss, shown as follows:

\vspace{-4mm}
\begin{align}
	\mathcal{L}_{\operatorname{KD}} 
	&= \alpha \cdot \mathcal{L}_{\operatorname{CE}} + \beta  \cdot\mathcal{L}_{\operatorname{KL}} \nonumber \\
	&= \alpha  \cdot\operatorname{CE}(\mathbf{q_{\tau=1}, \mathbf{y}}) + \beta  \cdot \tau^2  \cdot \operatorname \nonumber {KL}(\mathbf{p}_\tau, \mathbf{q}_\tau) 
\end{align}

where $\alpha$ and $\beta$ are hyper-parameters to balance the weight of the two losses. The reason for multiplying $\operatorname{KL}$ by  $\tau^2$ is to equalize the gradients of the two losses.

\subsection{Sharpness as a Unified Metric}
Recently, a few research works have found that a fixed temperature is not always optimal for the KD process. This issue has been investigated from two perspectives. From the model's perspective, tuning the temperature might alleviate the capability gap between the teacher and the student \cite{jafari2021annealing}, \cite{yuan2023student}. From the task's perspective, tuning the temperature might help set a proper task difficulty level  \cite{li2023curriculum}. Therefore, a comprehensive temperature regulation mechanism should ideally take into account both perspectives. 

Different from previous works, we propose a unified metric that has the potential to cover both issues. In particular, notice that the shape of a model's output logits represents its confidence in its prediction, which is influenced by both the prediction capability and the task difficulty. Therefore, we assume that a proper alignment of the output shapes of the teacher and the student might be able to mitigate problems in both perspectives. 
Note that a shared temperature can reduce the shape difference between the logits, but is not able to eliminate the difference. To tackle this issue, we first quantify the smoothness of output using the $\operatorname{logsumexp}$ function,  denoted as \textbf{sharpness}. Suppose the logit $\mathbf{z} = [z_1, z_2, ..., z_K]$, sharpness is defined as follows:

\vspace{-4mm}
\begin{align}
    \textit{sharpness}(\mathbf{z}) = \log \sum_{i=1}^K e^{z_i}
\end{align}

\begin{figure}[t]
    \centering
    \includegraphics[width=\linewidth]{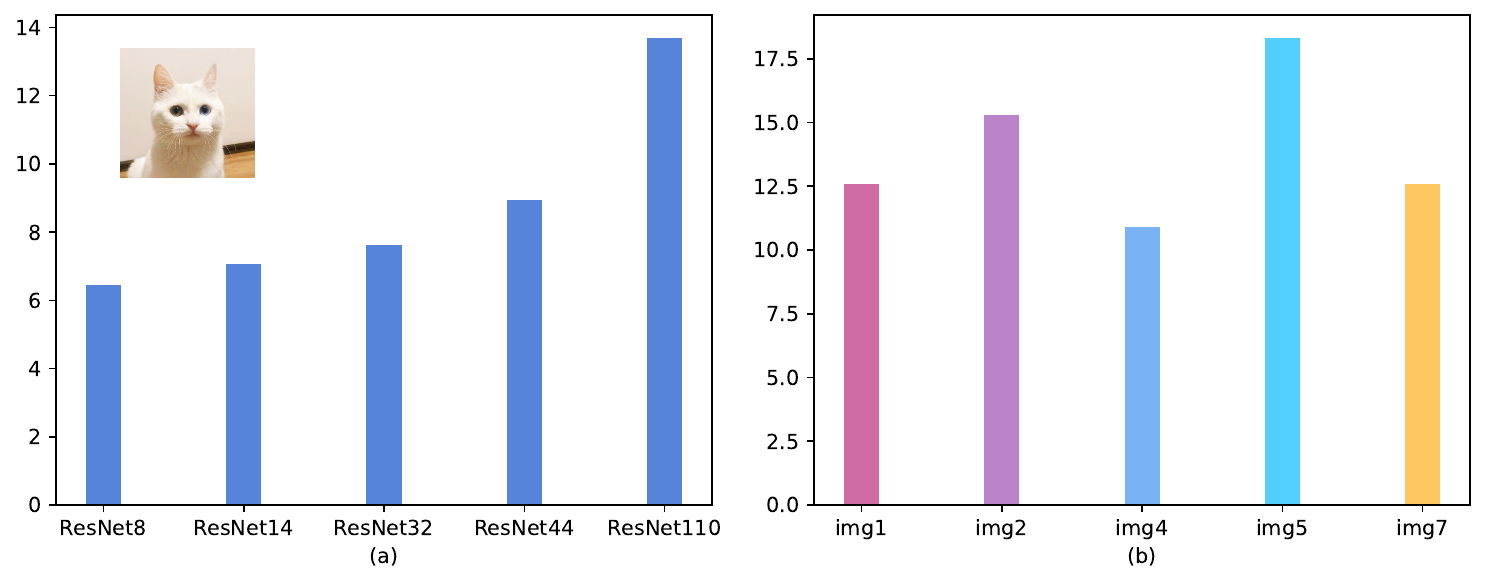}
    \caption{Sharpness values from (a) the same sample with different models, and (b) The same model with various samples. }
    \label{fig:sharpness}
\end{figure}

A high sharpness value indicates that the network's output logit has less smoothness, whereas a low sharpness value suggests greater smoothness. Therefore, we use sharpness to represent the smoothness of the logits, which is influenced by both the model's prediction capability and the sample's task difficulty, as shown in Figure~\ref{fig:sharpness}. As an example, networks of different prediction capabilities exhibit significantly varied sharpness, as shown in Figure~\ref{fig:sharpness} (a). A capable teacher's output is sharper since it is more confidence in its predictions. Within the same network model, different samples can exhibit varying levels of prediction difficulty. This variation consequently impacts the sharpness of the final output logits as well, as demonstrated in Figure 2 (b).

\subsection{Dynamic Temperature Knowledge Distillation}
\label{sec:dtkd}

Since a shared temperature cannot equalize the smoothness of two groups of logits, a natural idea is that the logits with greater sharpness (typically the output of the teacher network) should have a higher temperature, while the logits with lesser sharpness (typically the output of the student network) should have a lower temperature. Therefore our goal is to simultaneously regulate the temperatures of both the teacher and the student, such that the logits of the teacher are softened a bit more while the logits of the student are softened a bit less, and their levels of sharpness are equalized. We start from a common fixed temperature $\tau$ and set the teacher's temperature as $\tau + \delta$ while the student's temperature as $\tau - \delta$. We can find such a proper $\delta$ by minimizing the difference in sharpness between the teacher output $\mathbf u$ at temperature $\tau + \delta$ and the student output $\mathbf v$ at temperature $\tau - \delta$. That is, to find the proper $\delta$ such that:

\vspace{-4mm}
\begin{align}
    \arg \min _ {\delta} \left | \textit{sharpness}(\frac{\mathbf{u}}{\tau + \delta}) -  \textit{sharpness}(\frac{\mathbf{v}}{\tau - \delta}) \right|
    \label{formula:argmin}
\end{align}

Since the difference is a convex function, there is a unique $\delta$. In particular, we have the following proposition:

\begin{proposition}
\label{prop:inequality}
Assuming $\mathbf{u}$ and $\mathbf{v}$ are vectors in $\mathbb{R}^{n}$, and $\tau_1$, $\tau_2$ are two non-zero scalar real numbers, then we can derive

\begin{equation}
0 \le \left |\operatorname{logsumexp}(\frac{\mathbf u}{\tau_1} )-\operatorname{logsumexp}( \frac{\mathbf v}{\tau_2})\right| \le \left |\frac {\mathbf u}{\tau_1} - \frac{\mathbf v}{\tau_2} \right |_{\max}     \nonumber
\end{equation}
\end{proposition}

According to Proposition~\ref{prop:inequality}, to minimize the difference of formula~(\ref{formula:argmin}), we let $\tau_1=\tau + \delta$ and $\tau_2=\tau - \delta$, and set:

\begin{equation}
    \begin{aligned}
        \left |\frac {\mathbf u}{\tau + \delta} - \frac{\mathbf v}{\tau - \delta} \right |_{\max} = 0 \\
    \end{aligned}
\end{equation}

Let $|\mathbf u|_{\max} = x, |\mathbf v|_{\max} = y$,  we can get:

\vspace{-2mm}
\begin{equation}
\delta = \frac{x - y}{x + y} \ \tau 
\end{equation}

Finally, we can determine that the temperatures for the teacher and the student are:
$\tau + \delta = \frac{2x}{x+y}\ \tau$, $\tau - \delta = \frac{2y}{x+y}\ \tau$.

The loss function of DTKD is:

\vspace{-4mm}
\begin{align}
    \mathcal{L}_{\operatorname{DTKD}} =\frac{1}{N} \sum_{i=1}^N T_{tea} \cdot T_{stu} \cdot \operatorname {KL}(\mathbf{p}_{T_{tea}}, \mathbf{q}_{T_{stu}}) \nonumber 
\end{align}

where $T_{tea}=\frac{2x}{x+y}\ \tau$ and $T_{stu}=\frac{2y}{x+y}\ \tau$.
Numerous studies have shown that using a fixed temperature for knowledge distillation is effective in most cases. Therefore, in addition to DTKD, we also incorporate fixed-temperature KD. In practical applications, we assign it a relatively small loss weight. 

Finally, the complete loss function is:

\vspace{-4mm}
\begin{align}
    \mathcal{L}_{KD} 
	=  \alpha \cdot \mathcal{L}_{\operatorname{DTKD}} + \beta  \cdot\mathcal{L}_{\operatorname{KL}}  + \gamma \cdot \mathcal{L}_{\operatorname{CE}}  \nonumber
\end{align}

where $\alpha$, $\beta$, $\gamma$ are weight parameters for the three loss components respectively.

Algorithm \ref{alg:algorithmDTKD} provides the pseudo-code of DTKD in a PyTorch-like \cite{paszke2019pytorch} style which does not require any additional training costs or any additional modules.

\begin{algorithm}[t]
    \caption{Pseudo code of DTKD in a PyTorch-like style.}
    \label{alg:algorithmDTKD}
    \footnotesize
    \begin{alltt}
    \color{ForestGreen}
# l_stu: student output logits
# l_tea: teacher output logits
# l_stu_mx: maximum of student logits
# l_tea_mx: maximum of teacher logits
# T_stu: temperature for student in DTKD
# T_tea: temperature for teacher in DTKD
# p_stu: final prediction for student
# p_tea: final prediction for teacher
# alpha: hyper-parameters for DTKD
\end{alltt}

\begin{alltt}
\color{ForestGreen}# calculate the max of each sample's logit\color{Black}
l_stu_mx, _ = l_stu.max(dim=1, keepdim=True)
l_tea_mx, _ = l_tea.max(dim=1, keepdim=True)

\color{ForestGreen}# calculate the student and teacher network 
\color{ForestGreen}# temperature of each sample's logit \color{Black}
T_stu = 2 * l_stu_mx / (l_tea_mx+l_stu_mx)*T
T_tea = 2 * l_tea_mx / (l_tea_mx+l_stu_mx)*T
    
p_stu = F.softmax(l_student / T_stu)
p_tea = F.softmax(l_teacher / T_tea)
    
\color{ForestGreen}# DTKD \color{Black}
loss = kl_div(log(p_stu), p_tea)
dtkd_loss = alpha * loss * T_tea * T_stu
    \end{alltt}
\end{algorithm}

\subsection{Effectiveness of DTKD}

\begin{figure}[t]
    \centering
    \includegraphics[width=\linewidth]{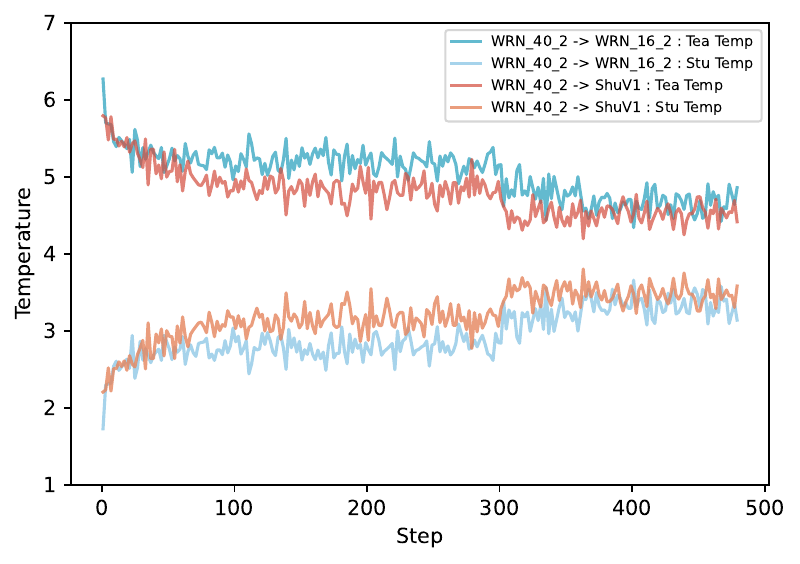}
    \caption{Teacher and student temperatures of DTKD over time.}
    \label{fig:temp_example}
\end{figure}

Notice that if our goal is to simply eliminate the difference of sharpness between the outputs of the teacher and the student, then there are infinitely many pairs of temperatures we can use. For the extreme cases, if we set the temperatures to positive infinite (or near zero), both shapes would be extremely flat (or sharp). However, our assumption suggests that we determine a proper $\delta$ that can help us find a moderate common ground for the smoothness of both the teacher and the student. In particular, during each iteration, $\delta$ is determined by both the reference temperature $\tau$ and the logits of the two models. As illustrated in Figure~\ref{fig:temp_example}, with DTKD, the temperatures of each pair of teacher and student are getting closer to each other during training. This is reasonable as the overall training difficulty should gradually increase. As the teacher's temperature decreases, the labels become ``harder''. This makes the task harder for the student because it gets less information of incorrect classes. From the student's point of view, as the student becomes more capable, increasing the temperature can continue providing rich information of incorrect classes to the teacher, and prevent the student from becoming too confident too quickly. Also note that for the same teacher (WRN\_40\_2) and different students (WRN\_16\_2 and ShuV1), the temperatures are adapted differently, indicating that DTKD also takes care of the capability gap of different pairs of models. In the next section, extensive experiments further demonstrate the effectiveness of our method.

\begin{table*}[ht]
    \centering
    \begin{tabular}{cc|cccccc}
        \toprule
        \multirow{4}{*}{
            \begin{tabular}[c]{@{}c@{}}distillation \\ manner\end{tabular}
        }   & teacher   & ResNet56 & ResNet110 & ResNet32$\times$4 & WRN-40-2 & WRN-40-2 & VGG13 \\
            &           & 72.34    & 74.31     & 79.42             & 75.61    & 75.61    & 74.64 \\
            & student   & ResNet20 & ResNet32  & ResNet8$\times$4  & WRN-16-2 & WRN-40-1 & VGG8 \\
            &           & 69.06    & 71.14     & 72.50             & 73.26    & 71.98    & 70.36 \\
        \toprule
        \multirow{5}{*}{features}   
            & FitNet    & 69.21 & 71.06 & 73.50 & 73.58 & 72.24 & 71.02 \\
            & RKD       & 69.61 & 71.82 & 71.90 & 73.35 & 72.22 & 71.48 \\
            & CRD       & 71.16 & 73.48 & 75.51 & 75.48 & 74.14 & 73.94 \\
            & OFD       & 70.98 & 73.23 & 74.95 & 75.24 & 74.33 & 73.95 \\
            & ReviewKD  & 71.89 & 73.89 & 75.63 & 76.12 & \textbf{75.09} & \textbf{74.84} \\
        \midrule
        \multirow{4}{*}{logits} 
            & KD                    & 70.66          & 73.08            & 73.33          & 74.92          & 73.54 & 72.98 \\
            & DKD                   & 71.97          & \textbf{74.11}   & \textbf{76.32} & \textbf{76.24} & 74.81 & 74.68 \\
            & $\text{NormKD}^{*}$   & 71.40          & 73.79            & 76.26          & 76.06          & 74.39 & 74.33 \\
            & \textbf{DTKD}         & \textbf{72.05} & 74.07            & 76.16          & 75.81          & 74.30 & 74.12 \\
        \midrule
            & $\Delta$ & \textcolor{ForestGreen}{+1.39} & \textcolor{ForestGreen}{+0.99} & \textcolor{ForestGreen}{+2.83} & \textcolor{ForestGreen}{+0.89} & \textcolor{ForestGreen}{+0.76} & \textcolor{ForestGreen}{+1.14} \\
        \bottomrule
    \end{tabular}
    \caption{\textbf{Results on the CIFAR-100 validation.} Teachers and students share the \textbf{same} architectures. $\Delta$ represents the performance improvement over the vanilla KD. $\text{NormKD}^{*}$: the statistics are reproduced by ourselves. All results are calculated by the average over 5 trials.}
    \label{tab:dtkd_same}
\end{table*}

\section{Experiments}

\subsection{Datasets and Settings} 
\begin{itemize}
\item  \textbf{CIFAR-100} \cite{krizhevsky2009learning} is a widely used dataset in deep learning, consisting of 60,000 color images of size 32x32, distributed across 100 classes, with 600 images per class. Each class has 500 training images and 100 test images, totaling 50,000 training images and 10,000 test images.

\item \textbf{ImageNet} \cite{russakovsky2015imagenet} is a well-known, large-scale classification dataset that covers 1,000 object classes and includes over 1.2 million training images and 50,000 validation images. Due to varying image resolutions, they are resized to 224 × 224 during preprocessing.

\item We selected neural networks with different architectures to serve as teacher or student models, including ResNet \cite{he2016deep}, WideResNet \cite{zagoruyko2016wide}, VGGNet \cite{simonyan2014very}, ShuffleNet-V1 \cite{zhang2018shufflenet} / V2 \cite{ma2018shufflenet}, and MobileNetV2 \cite{sandler2018mobilenetv2}.

\item A series of classical KD methods have been chosen for performance comparison. The ones based on Feature distillation are : FitNet \cite{romero2014fitnets}, AT \cite{zagoruyko2016paying}, RKD \cite{park2019relational}, CRD \cite{tian2019contrastive}, OFD \cite{heo2019comprehensive}, ReviewKD \cite{chen2021distilling}; the ones based on logits distillation are KD\cite{hinton2015distilling}, DKD\cite{zhao2022decoupled}, NormKD\cite{chi2023normkd}.
\end{itemize}

\subsection{Main Results}

\noindent \textbf{CIFAR-100 classification.}
We discuss the results of experiments on CIFAR-100 to validate the performance of our DTKD approach. The validation accuracy is reported in Table \ref{tab:dtkd_same} and Table \ref{tab:dtkd_diff}. Table \ref{tab:dtkd_same} contains the results where teachers and students are of the same network architectures. Table \ref{tab:dtkd_diff} shows the results where teachers and students are from different architectures. Table \ref{tab:dtkd_dkd} displays the results of combining DTKD and DKD, {where we can see that DTKD alone already reaches about the same performance compared to DKD, and with the two methods combined, we achieved top performance.}

\begin{table}[h]
    \centering
    \begin{tabular}{c|ccc}
        \toprule
        Teacher & ResNet56 & ResNet32$\times$4 & VGG13 \\
                & 72.34    & 79.42             & 74.64 \\
        Student & ResNet20 & ResNet8$\times$4  & VGG8  \\
                & 69.06    & 72.50             & 70.36 \\
        \midrule
        KD                                          & 70.66          & 73.33         & 72.98 \\
        DKD                                         & 71.97          & 76.32         & 74.68 \\
        \textbf{DTKD}                               & 72.05          & 76.16         & 74.12 \\
        \textbf{$\text{DTKD}^{\mathbf +}$} + DKD    & \textbf{72.20} &\textbf{76.93} &\textbf{74.87}\\
        \bottomrule
    \end{tabular}
    \caption{\textbf{Results on the CIFAR-100 validation.} $\text{DTKD}^{\mathbf +}$ indicates that we have replaced the fixed temperature in DKD with temperatures generated by our DTKD.}
    \label{tab:dtkd_dkd}
\end{table}

\noindent \textbf{ImageNet classification.}
Tables \ref{tab:dtkd_imagenet_same} and \ref{tab:dtkd_imagenet_diff} report the top-1 and top-5 accuracy rates for image classification on ImageNet. Compared to KD, our DTKD method shows improvements in both top-1 and top-5 accuracy.

\begin{table*}[htb]
    \centering
    \begin{tabular}{cc|ccccc}
        \toprule
        \multirow{4}{*}{
            \begin{tabular}[c]{@{}c@{}}
                distillation \\
                manner
            \end{tabular}
        } & teacher & ResNet32$\times$4 & WRN-40-2      & VGG13         & ResNet50      & ResNet32$\times$4 \\
          &         & 79.42             & 75.61         & 74.64         & 79.34         & 79.42 \\
          & student & ShuffleNet-V1     & ShuffleNet-V1 & MobileNet-V2  & MobileNet-V2  & ShuffleNet-V2 \\
          &         & 70.50             & 70.50         & 64.60         & 64.60         & 71.82 \\
        \toprule
        \multirow{5}{*}{features}   
            & FitNet     & 73.59 & 73.73 & 64.14 & 63.16 & 73.54 \\
            & RKD        & 72.28 & 72.21 & 64.52 & 64.43 & 73.21 \\
            & CRD        & 75.11 & 76.05 & 69.73 & 69.11 & 75.65 \\
            & OFD        & 75.98 & 75.85 & 69.48 & 69.04 & 76.82 \\
            & ReviewKD   & \textbf{77.45} & \textbf{77.14} & \textbf{70.37} & 69.89 & \textbf{77.78} \\
        \midrule
        \multirow{6}{*}{logits} 
            & KD                    & 74.07 & 74.83 & 67.37 & 67.35          & 74.45 \\
            & DKD                   & 76.45 & 76.70 & 69.71 & \textbf{70.35} & 77.07 \\
            & $\text{NormKD}^{*}$   & 75.40 & 76.39 & 68.92 & 69.53          & 75.82 \\
            & \textbf{DTKD}         & 75.43 & 76.29 & 69.01 & 69.10          & 76.19 \\
        \midrule
            & $\Delta$ &  \textcolor{ForestGreen}{+1.36} & \textcolor{ForestGreen}{+1.46} & \textcolor{ForestGreen}{+1.64} & \textcolor{ForestGreen}{+1.75} & \textcolor{ForestGreen}{+1.74} \\
        \bottomrule
    \end{tabular}
    \caption{\textbf{Results on the CIFAR-100 validation.} Teachers and students posses  \textbf{different} architectures. $\Delta$ represents the performance improvement over the vanilla KD. $\text{NormKD}^{*}$: the statistics are reproduced by ourselves. All results are calculated by the average over 5 trials.}
    \label{tab:dtkd_diff}
\end{table*}

\begin{table*}[htb]
    \centering
    \begin{tabular}{ccc|cccc|cccc}
        \toprule
       \multicolumn{3}{c|}{Distillation Manner} & \multicolumn{4}{c|}{Features} & \multicolumn{3}{c}{Logits} \\
        \midrule
                   & Teacher & Student & AT     & OFD   & CRD   & ReviewKD       & KD    & DKD            & \textbf{DTKD} \\
        Top-1      & 73.31   & 69.75   & 70.69  & 70.81 & 71.17 & 71.61          & 70.66 & \textbf{71.70} &  71.52 \\
        Top-5      & 91.42   & 89.07   & 90.01  & 89.98 & 90.13 & \textbf{90.51} & 89.88 & 90.41          & 90.15 \\ 
        \bottomrule
    \end{tabular}
    \caption{\textbf{Top-1 and top-5 accuracy (\%) on the ImageNet validation.} We set \textbf{ResNet-34} as the teacher model and \textbf{ResNet-18} as the student model. All results are calculated by the average over 3 trials.}
    \label{tab:dtkd_imagenet_same}
\end{table*}

\begin{table*}[htb]
    \centering
    \begin{tabular}{ccc|cccc|cccc}
        \toprule
        \multicolumn{3}{c|}{Distillation Manner} & \multicolumn{4}{c|}{Features} & \multicolumn{3}{c}{Logits} \\
        \midrule
                   & Teacher & Student & AT    & OFD   & CRD   & ReviewKD       & KD    & DKD            & \textbf{DTKD} \\
        Top-1      & 76.16   & 68.87   & 69.56 & 71.25 & 71.37 & \textbf{72.56} & 68.58 & 72.05          & 71.82 \\
        Top-5      & 92.86   & 88.76   & 89.33 & 90.34 & 90.41 & 91.00          & 88.98 & \textbf{91.05} & 90.28 \\
        \bottomrule
    \end{tabular}
    \caption{\textbf{Top-1 and top-5 accuracy (\%) on the ImageNet validation.} We set \textbf{ResNet-50} as the teacher model and \textbf{MobileNet-V1} as the student model.  All results are calculated by the average over 3 trials.}
    \label{tab:dtkd_imagenet_diff}
\end{table*}

\subsection{Training details}

In terms of experimental setup, we fully followed the protocol of DKD. For CIFAR-100: for most student models, the initial learning rate is set to 0.05, while for ShuffleNet and MobileNet, the initial learning rate is set to 0.01. Additionally, for all experiments: the batch size is set to 64, the number of training epochs is set to 240, the learning rate is reduced by a factor of 10 at epochs 150, 180, and 210, and there is a 20-epoch warm-up period to accelerate convergence. 

For ImageNet: We train the models for 100 epochs. With a batch size of 512, the learning rate starts at 0.2 and is divided by 10 every 30 epochs. For the same architecture, we choose ResNet34 as the teacher model and ResNet18 as the student model, with both $\alpha$ and $\beta$ set to 0.5. In the case of heterogeneous architecture, ResNet-50 is selected as the teacher model and MobileNet-V1 is chosen as the student model, with $\alpha$ and $\beta$ set to 1 and 0.5, respectively.

In all experiments, we use the SGD optimizer with a momentum setting of 0.9. Weight decay is set to 5e-4 for CIFAR-100 and 1e-4 for ImageNet. The loss weight $\gamma$ for CE (Cross-Entropy) is always 1. For KD, DKD, and DTKD, the temperature settings on CIFAR-100 and ImageNet are 4 and 1, respectively. All cases are trained on NVIDIA RTX 3090 GPUs, using 1 GPU for CIFAR-100 and 8 for ImageNet. 

\subsection{Extensions}
\subsubsection{Feature transferability.}
We conducted experiments to assess the transferability of deep features, to validate whether our DTKD method can transfer more universally applicable knowledge. We used Res32x4 as the teacher model and Res8x4 as the student model trained on the CIFAR-100 and performed transfer learning on the STL10 \cite{coates2011analysis} dataset. The process is designed as follows: we froze all parameters except the classification module and then carried out a limited number of training epochs to update the parameters of the classification layer only. The results, reported in Table ~\ref{tab:stl10}, clearly show that our DTKD and DKD are on the same level regarding knowledge transferability, while NormKD fell short by a big margin.

\begin{table}[H]
    \centering
    \begin{tabular}{c|cccc}
        \toprule
                    & KD    & DKD   & NormKD & DTKD  \\
        \midrule
        CIFAR-100   & 73.33 & 76.32 & 76.26  & 76.16  \\    
        STL-10      & 71.11 & 71.81 & 59.76  & 71.69   \\
        \bottomrule
    \end{tabular}
    \caption{The accuracy results of transfer learning from CIFAR-100 to STL-10.}
    \label{tab:stl10}
\end{table}

\subsubsection{Training efficiency.}
\begin{figure}[h]
    \centering
    \includegraphics[width=\linewidth]{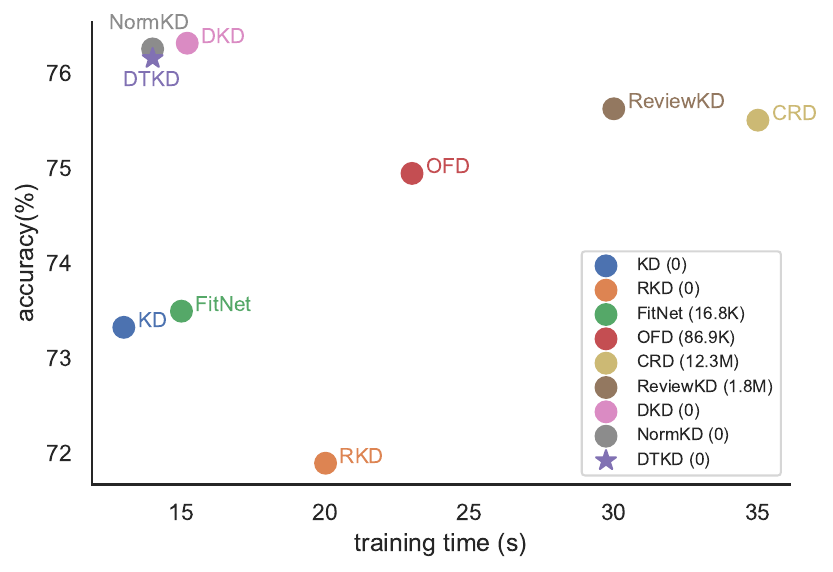}
    \caption{Training time (per epoch) vs. accuracy on CIFAR-100. We set \textbf{ResNet32$\times$4} as the teacher model and \textbf{ResNet8$\times$4} as the student model. The legend shows the extra parameters required in different methods.}
    \label{fig:speed}
\end{figure}

We assessed the training costs of state-of-the-art distillation methods, highlighting the high training efficiency of DTKD. As illustrated in Figure~\ref{fig:speed}, our DTKD's accuracy is among the top of all methods, while maintaining an ignorable additional cost with no extra parameters and a minimum amount of increased training time. Indeed, the only extra computation of DTKD is to calculate the maximal values of the logits of the teacher and the student.

\subsection{Robustness regarding TCKD and NCKD}

\begin{table}[h]
    \centering
    \begin{tabular}{c|cc|cc}
        \toprule
        Student & TCKD & NCKD & DKD & $\text{DTKD}^{\mathbf{+}}$ \\
        \midrule
        \multicolumn{5}{c}{\text{ResNet32$\times$4 as the teacher}} \\
        \midrule
        \multirow{3}{*}[-0.5ex]{\shortstack{ResNet8$\times$4 \\ 72.50}} & \checkmark & \checkmark & \textcolor{ForestGreen}{+1.13} & \textcolor{ForestGreen}{+2.63} \\
        & \checkmark & & \textcolor{red}{-3.87} & \textcolor{blue}{+0.25} \\
        & & \checkmark & \textcolor{ForestGreen}{+1.76} & \textcolor{ForestGreen}{+2.58} \\
        \midrule
        \multirow{3}{*}[-0.5ex]{\shortstack{ShuffleNet-V1 \\ 70.50}} & \checkmark & \checkmark & \textcolor{ForestGreen}{+3.79} & \textcolor{ForestGreen}{+3.84} \\
        & \checkmark & & \textcolor{red}{+0.02} & \textcolor{blue}{+1.42} \\
        & & \checkmark & \textcolor{ForestGreen}{+4.41} & \textcolor{ForestGreen}{+4.53} \\
        \midrule             
        \multicolumn{5}{c}{\text{WRN-40-2 as the teacher}} \\
        \midrule
        \multirow{3}{*}[-0.5ex]{\shortstack{WRN-16-2 \\ 73.26}} & \checkmark & \checkmark & \textcolor{ForestGreen}{+1.70} & \textcolor{ForestGreen}{+2.45} \\
        & \checkmark & & \textcolor{red}{-2.30} & \textcolor{blue}{+0.07} \\
        & & \checkmark & \textcolor{ForestGreen}{+1.50} & \textcolor{ForestGreen}{+2.50} \\
        \midrule
        \multirow{3}{*}[-0.5ex]{\shortstack{ShuffleNet-V1 \\ 70.50}} & \checkmark & \checkmark & \textcolor{ForestGreen}{+4.42} & \textcolor{ForestGreen}{+4.84} \\
        & \checkmark & & \textcolor{red}{+0.12} & \textcolor{blue}{+1.47} \\
        & & \checkmark & \textcolor{ForestGreen}{+4.62} & \textcolor{ForestGreen}{+4.91} \\                   
        \bottomrule
    \end{tabular}
    \caption{Accuracy(\%) on the CIFAR-100 validation set. $\text{DTKD}^{\mathbf{+}}$ indicates that we have replaced the fixed temperature in DKD with temperatures generated by our DTKD.}
    \label{tab:dtkd_tckd_nckd}
\end{table}

 We individually studied the influence of DTKD on TCKD and NCKD  \cite{zhao2022decoupled}. In particular, we choose CIFAR-100 \cite{krizhevsky2009learning} as the dataset, ResNet \cite{he2016deep}, WideResNet (WRN) \cite{zagoruyko2016wide}, and ShuffleNet \cite{zhang2018shufflenet} are selected as training models. We tested both homogeneous and heterogeneous model architectures for the teacher-student pairs. The experimental results are reported in Tables~\ref{tab:dtkd_tckd_nckd}, where the accuracy of each baseline model is shown below the model name of column ``Student''.  For each teacher-student pair, we conduct two groups of experiments. The group corresponding to the column ``DKD'' in the table shows the results of training by DKD with fixed temperature $4$. The group corresponding to the column ``$\text{DTKD}^{\mathbf +}$'' shows the results of training with DKD using temperatures generated by DTKD with reference temperature $4$. In each group, we trained on either TCKD only, or NCKD only, or with both TCKD and NCKD. The weight of each loss is set at 1.0 (including the default cross-entropy loss). The table shows that solely distilling TCKD with fixed temperature can severely inhibit the distillation effect, while with DTKD can eliminate this defect. We suspect the reason is that the collaboration between the teacher and the student is crucial when only distilling the knowledge of the target classes. This experiment indicates that when dealing with different styles of knowledge distillation, DTKD has the potential to increase the robustness of the KD process.

\subsection{Visualizations}
\begin{figure}[htp!]
    \centering
    \includegraphics[width=\linewidth]{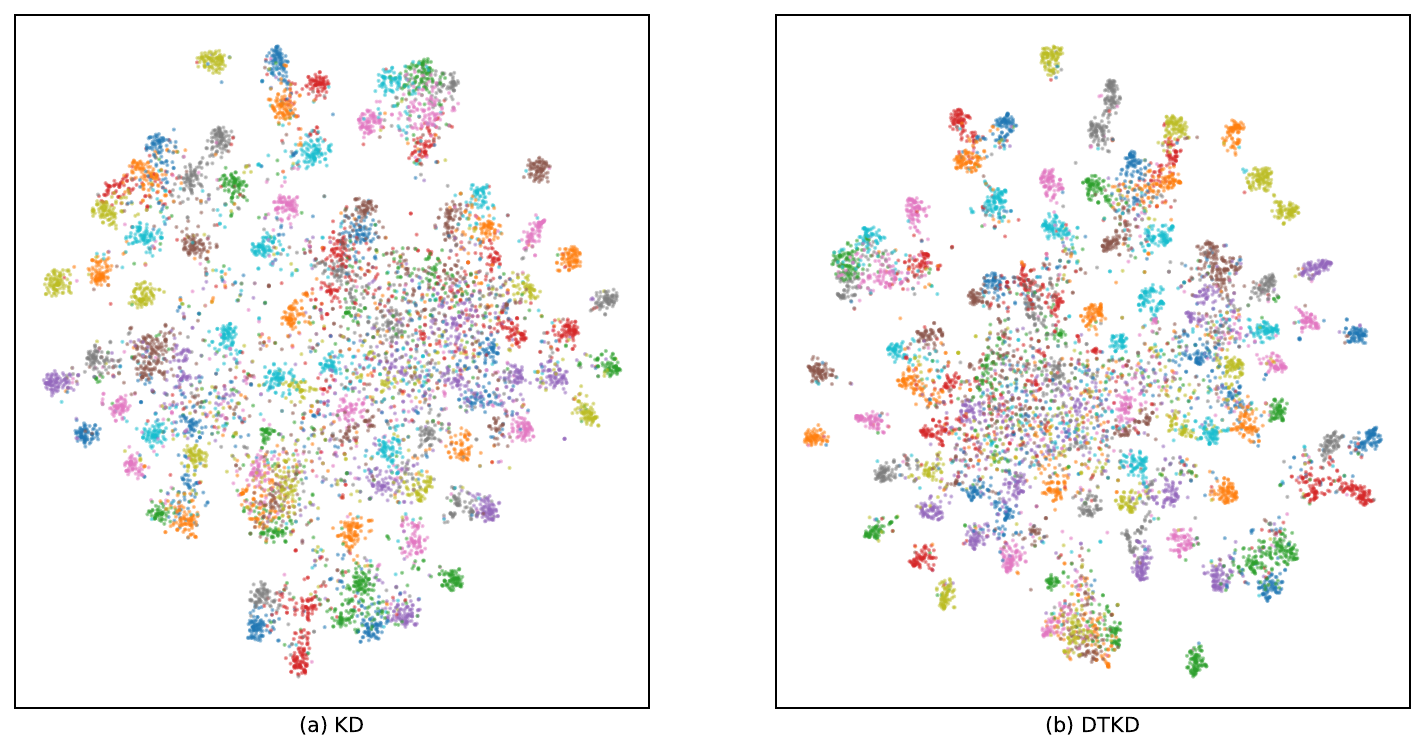}
    \caption{t-SNE of features learned by KD (left) and DTKD (right).}
    \label{fig:t_SNE}
\end{figure}

We present the visualization by t-SNE with the setting of ResNet32$\times$4 as teacher and ResNet8$\times$4 as student on CIFAR-100. Figure~\ref{fig:t_SNE} displays the visual comparison between KD and DTKD. The results indicate that our method exhibits higher separability compared to KD.

\section{Discussion and Conclusion}
In this paper, we conduct a series of studies on Dynamic Temperature Knowledge Distillation (DTKD) in an effort to mitigate the issues brought about by the traditional fixed temperature KD. We first introduce a metric to quantify the degree of logit softening, namely \textbf{sharpness}. Based on minimizing the difference in sharpness between the teacher and the student, we propose DTKD. It collaboratively finds more suitable temperatures for both the teacher and the student in each round of training. Moreover, our method significantly enhances the effectiveness of KD, especially when the student's learning ability is relatively weak. The DTKD approach is notably simple, requiring almost no additional overhead, and can be effectively integrated with other methods like DKD. Extensive experiments demonstrate the effectiveness of our method.

\section{Limitation and Future Work}
Like vanilla KD, we still need to set up a reference temperature, which relies on empirical experience. In future work, we plan to further investigate the selection of the reference temperature adaptively. Another direction to be explored is to study the influence of capability gap and task difficulty in dynamic temperature regulation respectively, and further optimize DTKD.

\newpage
\bibliographystyle{named}
\bibliography{ijcai24}

\begin{thebibliography}{}

\bibitem[\protect\citeauthoryear{Chen \bgroup \em et al.\egroup }{2021}]{chen2021distilling}
Pengguang Chen, Shu Liu, Hengshuang Zhao, and Jiaya Jia.
\newblock Distilling knowledge via knowledge review.
\newblock In {\em Proceedings of the IEEE/CVF Conference on Computer Vision and Pattern Recognition}, pages 5008--5017, 2021.

\bibitem[\protect\citeauthoryear{Chen \bgroup \em et al.\egroup }{2022}]{chen2022knowledge}
Defang Chen, Jian-Ping Mei, Hailin Zhang, Can Wang, Yan Feng, and Chun Chen.
\newblock Knowledge distillation with the reused teacher classifier.
\newblock In {\em Proceedings of the IEEE/CVF conference on computer vision and pattern recognition}, pages 11933--11942, 2022.

\bibitem[\protect\citeauthoryear{Chi \bgroup \em et al.\egroup }{2023}]{chi2023normkd}
Zhihao Chi, Tu~Zheng, Hengjia Li, Zheng Yang, Boxi Wu, Binbin Lin, and Deng Cai.
\newblock Normkd: Normalized logits for knowledge distillation.
\newblock {\em arXiv preprint arXiv:2308.00520}, 2023.

\bibitem[\protect\citeauthoryear{Coates \bgroup \em et al.\egroup }{2011}]{coates2011analysis}
Adam Coates, Andrew Ng, and Honglak Lee.
\newblock An analysis of single-layer networks in unsupervised feature learning.
\newblock In {\em Proceedings of the fourteenth international conference on artificial intelligence and statistics}, pages 215--223. JMLR Workshop and Conference Proceedings, 2011.

\bibitem[\protect\citeauthoryear{He \bgroup \em et al.\egroup }{2016}]{he2016deep}
Kaiming He, Xiangyu Zhang, Shaoqing Ren, and Jian Sun.
\newblock Deep residual learning for image recognition.
\newblock In {\em Proceedings of the IEEE conference on computer vision and pattern recognition}, pages 770--778, 2016.

\bibitem[\protect\citeauthoryear{Heo \bgroup \em et al.\egroup }{2019}]{heo2019comprehensive}
Byeongho Heo, Jeesoo Kim, Sangdoo Yun, Hyojin Park, Nojun Kwak, and Jin~Young Choi.
\newblock A comprehensive overhaul of feature distillation.
\newblock In {\em Proceedings of the IEEE/CVF International Conference on Computer Vision}, pages 1921--1930, 2019.

\bibitem[\protect\citeauthoryear{Hinton \bgroup \em et al.\egroup }{2015}]{hinton2015distilling}
Geoffrey Hinton, Oriol Vinyals, and Jeff Dean.
\newblock Distilling the knowledge in a neural network.
\newblock {\em arXiv preprint arXiv:1503.02531}, 2015.

\bibitem[\protect\citeauthoryear{Jafari \bgroup \em et al.\egroup }{2021}]{jafari2021annealing}
Aref Jafari, Mehdi Rezagholizadeh, Pranav Sharma, and Ali Ghodsi.
\newblock Annealing knowledge distillation.
\newblock {\em arXiv preprint arXiv:2104.07163}, 2021.

\bibitem[\protect\citeauthoryear{Krizhevsky \bgroup \em et al.\egroup }{2009}]{krizhevsky2009learning}
Alex Krizhevsky, Geoffrey Hinton, et~al.
\newblock Learning multiple layers of features from tiny images.
\newblock 2009.

\bibitem[\protect\citeauthoryear{Li \bgroup \em et al.\egroup }{2017}]{li2017learning}
Yuncheng Li, Jianchao Yang, Yale Song, Liangliang Cao, Jiebo Luo, and Li-Jia Li.
\newblock Learning from noisy labels with distillation.
\newblock In {\em Proceedings of the IEEE international conference on computer vision}, pages 1910--1918, 2017.

\bibitem[\protect\citeauthoryear{Li \bgroup \em et al.\egroup }{2022}]{li2022asymmetric}
Xin-Chun Li, Wen-Shu Fan, Shaoming Song, Yinchuan Li, Shao Yunfeng, De-Chuan Zhan, et~al.
\newblock Asymmetric temperature scaling makes larger networks teach well again.
\newblock {\em Advances in Neural Information Processing Systems}, 35:3830--3842, 2022.

\bibitem[\protect\citeauthoryear{Li \bgroup \em et al.\egroup }{2023}]{li2023curriculum}
Zheng Li, Xiang Li, Lingfeng Yang, Borui Zhao, Renjie Song, Lei Luo, Jun Li, and Jian Yang.
\newblock Curriculum temperature for knowledge distillation.
\newblock In {\em Proceedings of the AAAI Conference on Artificial Intelligence}, volume~37, pages 1504--1512, 2023.

\bibitem[\protect\citeauthoryear{Liu \bgroup \em et al.\egroup }{2022}]{liu2022meta}
Jihao Liu, Boxiao Liu, Hongsheng Li, and Yu~Liu.
\newblock Meta knowledge distillation.
\newblock {\em arXiv preprint arXiv:2202.07940}, 2022.

\bibitem[\protect\citeauthoryear{Ma \bgroup \em et al.\egroup }{2018}]{ma2018shufflenet}
Ningning Ma, Xiangyu Zhang, Hai-Tao Zheng, and Jian Sun.
\newblock Shufflenet v2: Practical guidelines for efficient cnn architecture design.
\newblock In {\em Proceedings of the European conference on computer vision (ECCV)}, pages 116--131, 2018.

\bibitem[\protect\citeauthoryear{Mirzadeh \bgroup \em et al.\egroup }{2020}]{mirzadeh2020improved}
Seyed~Iman Mirzadeh, Mehrdad Farajtabar, Ang Li, Nir Levine, Akihiro Matsukawa, and Hassan Ghasemzadeh.
\newblock Improved knowledge distillation via teacher assistant.
\newblock In {\em Proceedings of the AAAI conference on artificial intelligence}, volume~34, pages 5191--5198, 2020.

\bibitem[\protect\citeauthoryear{Park \bgroup \em et al.\egroup }{2019}]{park2019relational}
Wonpyo Park, Dongju Kim, Yan Lu, and Minsu Cho.
\newblock Relational knowledge distillation.
\newblock In {\em Proceedings of the IEEE/CVF conference on computer vision and pattern recognition}, pages 3967--3976, 2019.

\bibitem[\protect\citeauthoryear{Paszke \bgroup \em et al.\egroup }{2019}]{paszke2019pytorch}
Adam Paszke, Sam Gross, Francisco Massa, Adam Lerer, James Bradbury, Gregory Chanan, Trevor Killeen, Zeming Lin, Natalia Gimelshein, Luca Antiga, et~al.
\newblock Pytorch: An imperative style, high-performance deep learning library.
\newblock {\em Advances in neural information processing systems}, 32, 2019.

\bibitem[\protect\citeauthoryear{Romero \bgroup \em et al.\egroup }{2014}]{romero2014fitnets}
Adriana Romero, Nicolas Ballas, Samira~Ebrahimi Kahou, Antoine Chassang, Carlo Gatta, and Yoshua Bengio.
\newblock Fitnets: Hints for thin deep nets.
\newblock {\em arXiv preprint arXiv:1412.6550}, 2014.

\bibitem[\protect\citeauthoryear{Russakovsky \bgroup \em et al.\egroup }{2015}]{russakovsky2015imagenet}
Olga Russakovsky, Jia Deng, Hao Su, Jonathan Krause, Sanjeev Satheesh, Sean Ma, Zhiheng Huang, Andrej Karpathy, Aditya Khosla, Michael Bernstein, et~al.
\newblock Imagenet large scale visual recognition challenge.
\newblock {\em International journal of computer vision}, 115:211--252, 2015.

\bibitem[\protect\citeauthoryear{Sandler \bgroup \em et al.\egroup }{2018}]{sandler2018mobilenetv2}
Mark Sandler, Andrew Howard, Menglong Zhu, Andrey Zhmoginov, and Liang-Chieh Chen.
\newblock Mobilenetv2: Inverted residuals and linear bottlenecks.
\newblock In {\em Proceedings of the IEEE conference on computer vision and pattern recognition}, pages 4510--4520, 2018.

\bibitem[\protect\citeauthoryear{Simonyan and Zisserman}{2014}]{simonyan2014very}
Karen Simonyan and Andrew Zisserman.
\newblock Very deep convolutional networks for large-scale image recognition.
\newblock {\em arXiv preprint arXiv:1409.1556}, 2014.

\bibitem[\protect\citeauthoryear{Tang \bgroup \em et al.\egroup }{2020}]{tang2020understanding}
Jiaxi Tang, Rakesh Shivanna, Zhe Zhao, Dong Lin, Anima Singh, Ed~H Chi, and Sagar Jain.
\newblock Understanding and improving knowledge distillation.
\newblock {\em arXiv preprint arXiv:2002.03532}, 2020.

\bibitem[\protect\citeauthoryear{Tian \bgroup \em et al.\egroup }{2019}]{tian2019contrastive}
Yonglong Tian, Dilip Krishnan, and Phillip Isola.
\newblock Contrastive representation distillation.
\newblock {\em arXiv preprint arXiv:1910.10699}, 2019.

\bibitem[\protect\citeauthoryear{Yuan \bgroup \em et al.\egroup }{2020}]{yuan2020revisiting}
Li~Yuan, Francis~EH Tay, Guilin Li, Tao Wang, and Jiashi Feng.
\newblock Revisiting knowledge distillation via label smoothing regularization.
\newblock In {\em Proceedings of the IEEE/CVF Conference on Computer Vision and Pattern Recognition}, pages 3903--3911, 2020.

\bibitem[\protect\citeauthoryear{Yuan \bgroup \em et al.\egroup }{2023}]{yuan2023student}
Mengyang Yuan, Bo~Lang, and Fengnan Quan.
\newblock Student-friendly knowledge distillation.
\newblock {\em arXiv preprint arXiv:2305.10893}, 2023.

\bibitem[\protect\citeauthoryear{Zagoruyko and Komodakis}{2016a}]{zagoruyko2016paying}
Sergey Zagoruyko and Nikos Komodakis.
\newblock Paying more attention to attention: Improving the performance of convolutional neural networks via attention transfer.
\newblock {\em arXiv preprint arXiv:1612.03928}, 2016.

\bibitem[\protect\citeauthoryear{Zagoruyko and Komodakis}{2016b}]{zagoruyko2016wide}
Sergey Zagoruyko and Nikos Komodakis.
\newblock Wide residual networks.
\newblock {\em arXiv preprint arXiv:1605.07146}, 2016.

\bibitem[\protect\citeauthoryear{Zhang \bgroup \em et al.\egroup }{2018a}]{zhang2018shufflenet}
Xiangyu Zhang, Xinyu Zhou, Mengxiao Lin, and Jian Sun.
\newblock Shufflenet: An extremely efficient convolutional neural network for mobile devices.
\newblock In {\em Proceedings of the IEEE conference on computer vision and pattern recognition}, pages 6848--6856, 2018.

\bibitem[\protect\citeauthoryear{Zhang \bgroup \em et al.\egroup }{2018b}]{zhang2018deep}
Ying Zhang, Tao Xiang, Timothy~M Hospedales, and Huchuan Lu.
\newblock Deep mutual learning.
\newblock In {\em Proceedings of the IEEE conference on computer vision and pattern recognition}, pages 4320--4328, 2018.

\bibitem[\protect\citeauthoryear{Zhao \bgroup \em et al.\egroup }{2022}]{zhao2022decoupled}
Borui Zhao, Quan Cui, Renjie Song, Yiyu Qiu, and Jiajun Liang.
\newblock Decoupled knowledge distillation.
\newblock In {\em Proceedings of the IEEE/CVF Conference on computer vision and pattern recognition}, pages 11953--11962, 2022.

\bibitem[\protect\citeauthoryear{Zhou \bgroup \em et al.\egroup }{2021}]{zhou2021rethinking}
Helong Zhou, Liangchen Song, Jiajie Chen, Ye~Zhou, Guoli Wang, Junsong Yuan, and Qian Zhang.
\newblock Rethinking soft labels for knowledge distillation: A bias-variance tradeoff perspective.
\newblock {\em arXiv preprint arXiv:2102.00650}, 2021.

\end{thebibliography}

\clearpage
\appendix

\section{Detailed Explanations}
\begin{table*}[t]
    \centering
    \begin{tabular}{c|cc|c|c|c|c}
        \toprule
        Method      &Extra Training &Extra Module& Consistent smoothness & Synergy & Model Variety & Task Difficulty\\
        \midrule
        Annealing KD & \Checkmark &            & \XSolidBrush & \XSolidBrush  & \Checkmark    & \Checkmark \\
        MKD          & \Checkmark  & \Checkmark & \XSolidBrush & \Checkmark  & \Checkmark    & \Checkmark \\
        CTKD         &            & \Checkmark & \XSolidBrush & \XSolidBrush  & \Checkmark  & \Checkmark \\
        NormKD       &            &            & \Checkmark   & \XSolidBrush  & \XSolidBrush  & \XSolidBrush \\
        DTKD         &            &            & \Checkmark   & \Checkmark    & \Checkmark    & \Checkmark \\
        \bottomrule
    \end{tabular}
    \caption{Summary of important features of related works in dynamic temperature regulation.}
    \label{tab:compare}
\end{table*}

\subsection{Further Explanation of Table 1}
In the full paper, the details in Table~\ref{tab:compare} are not covered in detail due to lack of space. Here we present a comprehensive explanation. In Table~\ref{tab:compare}, we compared the previous temperature regulation methods with our method in five ways: (1) Whether additional resources are needed to assist the distillation, (2) Consistent smoothness, whether the method unifies smoothness for both peers, (3) Temperature synergy, which refers to whether the method takes into account the cooperation between the teacher and the student, (4) Model variety: whether the gap between various pairs of student and teacher are taken into account, and (5) Task difficulty, whether the method takes into account the task difficulty as training goes on. We summarize the table's content as follows according to the columns:

\begin{itemize}
    \item Extra Training: In Annealing-KD, the teacher model's outputs are modified at various temperatures and are used for training the student model iteratively. This leads to extra training across the whole dataset. In MKD and CTKD, extra training is required to train the added MLPs only (for temperature generation), and we do not count them as extra training. However, MKD is required to train on extra validation samples. No other method requires extra training in any way.
    \item Extra Module: In both MKD and CTKD,  extra MLP networks are used to predict the temperature parameters. No extra modules are employed in the other methods.
    \item Consistent smoothness: In Annealing-KD, only the teacher has applied a temperature, while the student's raw logits are paired with different saturated logits of the teacher, therefore the smoothness between the logits of teacher and student is not consistent. MKD and CTKD also do not require smoothness to be consistent. Both NormKD and our DTKD insist on smoothness consistency.
    \item Synergy: In Annealing-KD, since the student has no temperature parameter, therefore the method does not consider the collaborative temperatures of the peers. In MKD, the temperatures of both the student and the teacher are generated cooperatively by an MLP. In CTKD, only a shared temperature is computed which limits effective collaboration between the teacher and the student. In NormKD, since logits are normalized across the training process, the teacher's temperature is not affected by the student's outputs, therefore there is no collaboration. Our DTKD coordinates the temperatures of the teacher and the student by adjusting a $\delta$ that is shared in deriving their temperatures respectively.
    \item Model Variety: In Annealing-KD, the student model is selected based on different annealing iterations, therefore the variety of peered models is considered. MKD takes care of model variety by dynamically learning the temperatures based on student's performance. CTKD copes with model variety by predicting a proper shared temperature via adversarial training. In NormKD, a teacher outputs the same normalized logits regardless of different students. Instead, our DTKD adjusts temperatures differently according to the capability of different models.
    \item Task Difficulty: Since Annealing-KD fine-tunes the hard samples in the second training phase, it does take task difficulty into account. MKD dynamically learns the temperatures based on the student's performance, therefore also adjusts according to the task difficulty. CTKD explicitly adjusts task difficulty by employing curriculum learning. NormKD's teacher outputs the same normalized logits, therefore does not adjust according to task difficulty. DTKD regulates the temperatures of the models to be closer to each other, making the training task gradually more difficult for the student. 
\end{itemize}

\subsection{Proof of Proposition 1}
\label{app:appendix_1}
\begin{proposition}
\label{prop:inequality}
Assuming $\mathbf{u}$ and $\mathbf{v}$ are vectors in $\mathbb{R}^{n}$, and $\tau_1$, $\tau_2$ are two non-zero scalar real numbers, then we can derive

\begin{equation}
0 \le \left |\operatorname{logsumexp}(\frac{\mathbf u}{\tau_1} )-\operatorname{logsumexp}( \frac{\mathbf v}{\tau_2})\right| \le \left |\frac {\mathbf u}{\tau_1} - \frac{\mathbf v}{\tau_2} \right |_{\max}     \nonumber
\end{equation}
\end{proposition}

\begin{proof}
Let's return to the definition of logsumexp: for $\mathbf z = [z_1, z_2, ..., z_K]$, we have

\begin{align}
\operatorname{logsumexp}(\mathbf{z}) = \log \sum_{i=1}^K e^{z_i}
\end{align}

Assume $f(t) = \text{logsumexp}(t\cdot \mathbf{u} + (1-t)\cdot \mathbf{v})$. Then we can obtain : 

\begin{equation}
\left\{
    \begin{aligned}
    f(0) &= \operatorname{logsumexp}(\mathbf{v}) \\
    \\
    f(1) &= \operatorname{logsumexp}(\mathbf{u})
    \end{aligned}
\right.
\end{equation}

According to the mean value theorem, there exists $\varepsilon \in (0, 1)$, such that

\begin{equation}
    \begin{aligned}
    f'(\varepsilon) 
    &= \frac{f(1) - f(0)}{1 - 0}\\
    &= \operatorname{logsumexp}(\mathbf u) - \operatorname{logsumexp}(\mathbf v)
\end{aligned}
\end{equation}

Based on the definition of $\operatorname{logsumexp}$, then
\begin{equation}
    \resizebox{.91\linewidth}{!}{$
            \displaystyle
            \left |\operatorname{logsumexp}(\mathbf u)-\operatorname{logsumexp}(\mathbf v) \right |= \left|\frac{\sum_{i=1}^K e^{\varepsilon u_i+(1-\varepsilon) v_i}(u_i-v_i)}{\sum_{i=1}^K e^{\varepsilon u_i+(1-\varepsilon) v_i}}\right|
        $}
\end{equation}%

\begin{equation}
    \begin{aligned}
        & \leq \frac{\sum_{i=1}^K e^{\varepsilon u_i+(1-\varepsilon) v_i}|u_i-v_i|}{\sum_{i=1}^K e^{\varepsilon u_i+(1-\varepsilon) v_i}} \\ 
        & \leq \frac{\sum_{i=1}^K e^{\varepsilon u_i+(1-\varepsilon) v_i}|\mathbf u- \mathbf v|_{\infty}}{\sum_{i=1}^K e^{\varepsilon u_i+(1-\varepsilon) v_i}}\\
        & = |\mathbf u- \mathbf v|_{\infty} \\
        & = |\mathbf u- \mathbf v|_{\max}
    \end{aligned}
\end{equation}

Finally, we can get :
\begin{equation}
0 \le \left |\operatorname{logsumexp}(\frac{\mathbf u}{\tau_1} )-\operatorname{logsumexp}( \frac{\mathbf v}{\tau_2})\right| \le \left |\frac {\mathbf u}{\tau_1} - \frac{\mathbf v}{\tau_2} \right |_{\max}     \nonumber
\end{equation}
\end{proof}

\subsection{Implementation: Details in Feature Transferability}
In section \textbf{4.4 Extensions} of the full paper, transferability is assessed to validate whether our DTKD can transfer more universally applicable knowledge. Here more detailed experimental settings are provided. We selected ResNet-32$\times$4 as the teacher model and ResNet8-$\times$4 as the student model. We performed distillation with the following methods on the CIFAR-100 dataset: vanilla KD, DKD, NormKD, and our DTKD. After obtaining student models from these distillation methods, we conducted feature transferability experiments on the STL-10 Dataset. Specifically, we froze all parameters except for those in the classification layer and then modified the classifier so that the model's output could be correctly mapped to the new dataset's categories. The hyperparameters are: the batch size is set to $128$, the number of training epochs is set to $40$, the initial learning rate is set to $0.1$, and the learning rate is divided by 10 for every $10$ epochs. We chose SGD as the optimizer, with $0.9$ momentum and $0.0$ weight decay.

\subsection{More Effectiveness of DTKD}
Figure 1 of the full paper shows the effectiveness of DTKD when applying homogeneous distillation, i.e. teacher and student share the same network architecture (ResNet), illustrating the effectiveness of our method in bridging the capability gap between teacher and student models. Here we explored two more scenarios. 

\begin{figure}[h]
    \centering
    \includegraphics[width=\linewidth]{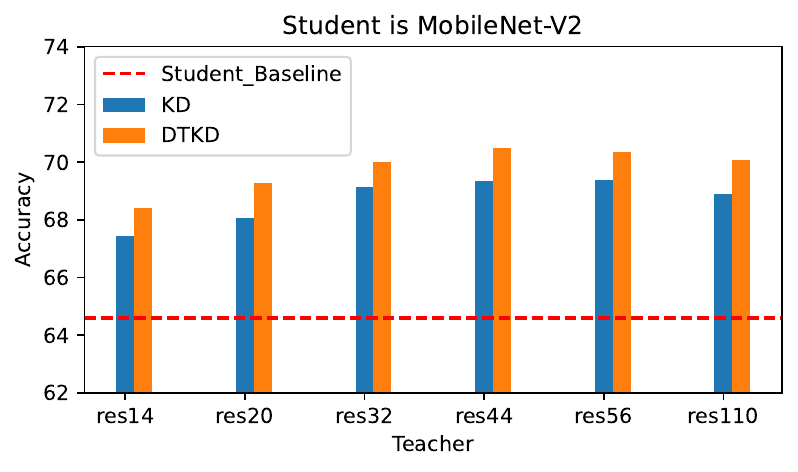}
    \caption{Heterogeneous distillation.}
    \label{fig:mobilenetv2}
\end{figure}

\begin{figure}[H]
    \centering
    \includegraphics[width=\linewidth]{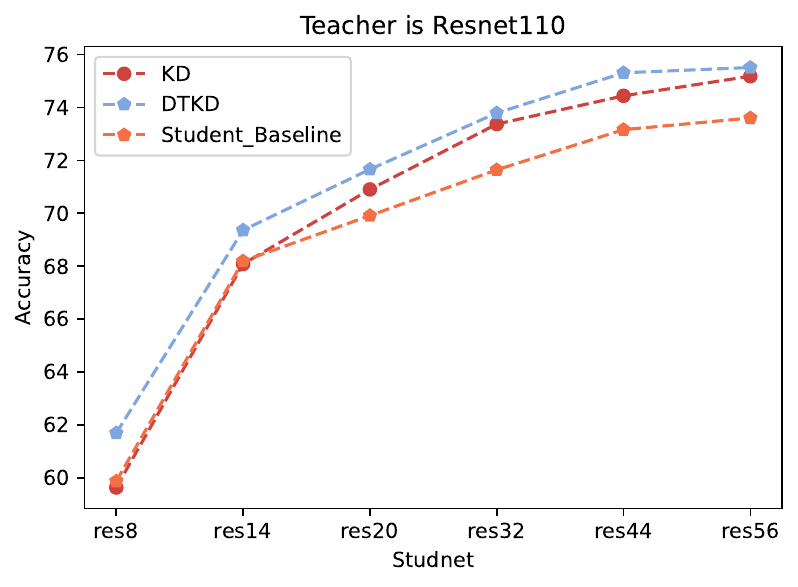}
    \caption{Paring the same teacher with various students.}
    \label{fig:diff_teacher}
\end{figure}

\begin{itemize}
    \item Firstly, as shown in Figure~\ref{fig:mobilenetv2}, we performed heterogeneous distillation, where ResNet is selected as the teacher models' architecture, and MobileNet-V2 is selected as the student model. for each concrete teacher model, DTKD performs better than vanilla KD. 
    \item Secondly, we fixed a highly capable teacher model (ResNet110), and employed it to distill knowledge into a series of student models with varying capacities (ResNet8, ResNet14, ResNet20, ResNet32, ResNet44, ResNet56). As depicted in Figure \ref{fig:diff_teacher}, DTKD demonstrates adaptability to various student-teacher pairs. Notably, even when the student model has a poor learning capacity, our method maintains commendable performance.
\end{itemize}

\begin{figure}[ht]
    \centering
    \includegraphics[width=\linewidth]{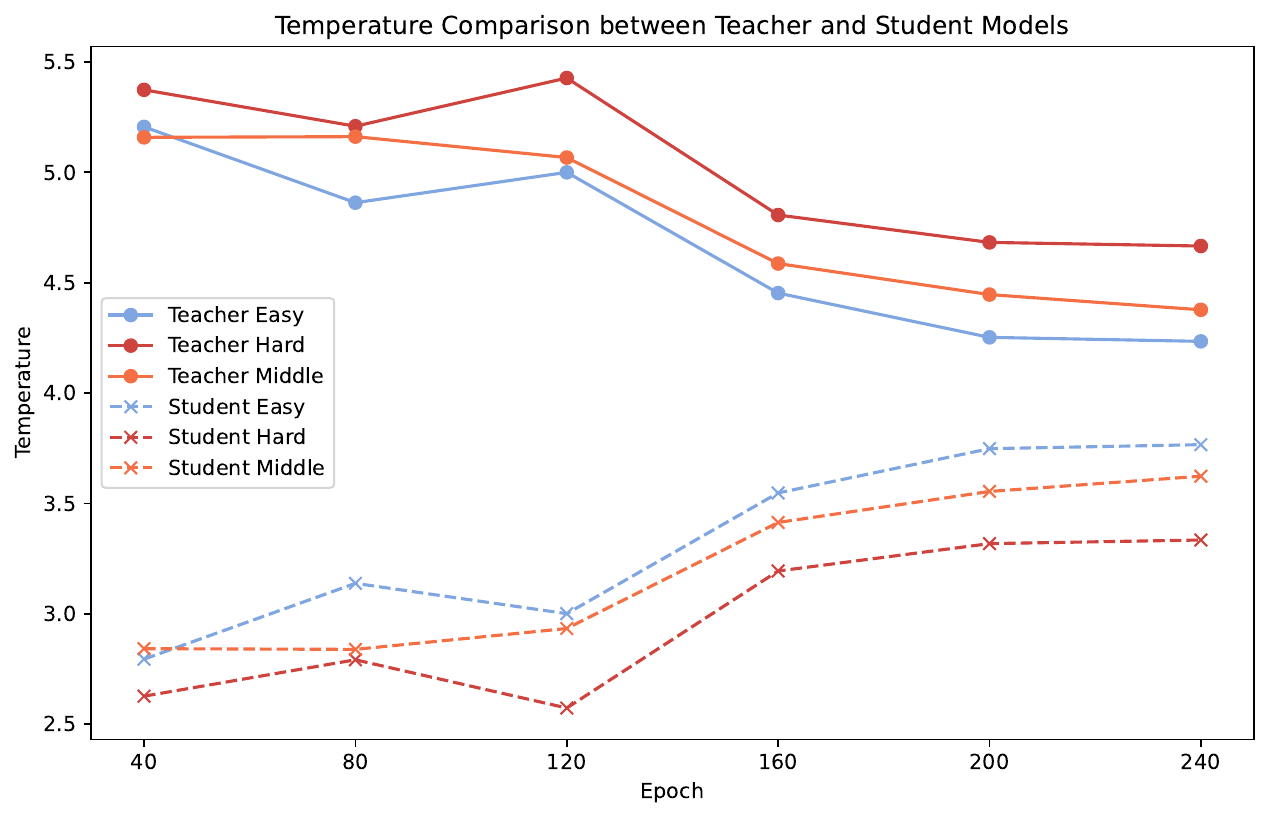}
    \caption{DTKD's temperature behavior for samples of different difficulties.}
    \label{fig:diff_sample_temp}
\end{figure}

Finally, Figure~\ref{fig:diff_sample_temp} shows that DTKD effectively adjusts the temperatures according to samples of different difficulty. In particular, according to the maximum value output by the teacher network, the first 30 (easy), middle 30 (middle), and last 30 (hard) of 1000 logit samples were selected. Then the temperatures of these samples at different epochs were calculated. As shown in the figure:

\begin{itemize}
    \item The teacher's temperature is lower for easy samples than for hard samples. This indicates that the teacher reinforces the correct learning patterns by providing ``harder'' labels. A harder label of an easy sample also effectively tells the student that fewer learning resources are required, allowing the student to focus on learning from the harder, softer labels where the additional information is more valuable.
    For harder samples, the teacher's temperature is higher, suggesting it is less confident and therefore provides softer labels, conveying more about the relative information of incorrect classes (the additional ``dark knowledge''). This is especially valuable for harder samples where the decision boundaries may not be as clear.
    \item The student's temperature is higher for easy samples, which might seem counterintuitive at first since higher temperatures typically soften the output distribution. However, as easy samples are more vulnerable to overfitting, a higher temperature effectively prevents the student from becoming overconfident on easy samples, ensuring it remains receptive to learning from the teacher. 
    Conversely, the student's temperature for harder samples is lower, which would typically increase the model's confidence in its predictions. In this context, it encourages the student to make more definitive predictions about harder samples, trying to push the student to learn these more challenging examples thoroughly.
\end{itemize}

To further validate the performance across the entire CIFAR-100 dataset, we categorized the test dataset by difficulty based on the teacher network's outputs. We sorted the test samples according to the maximum output of the teacher network, ranging from lowest to highest. The dataset was then divided into three groups: the top $33\%$ as \textbf{hard}, the bottom $33\%$ as \textbf{easy}, and the remaining $34\%$ as \textbf{medium} difficulty.  We subsequently evaluated the performance of both KD and DTKD on these sample groups with distinct difficulty levels, using accuracy as the key metric. As indicated in Table ~\ref{tab:hrad_mid_easy}, the DTKD method consistently outperforms KD in accuracy across all levels of difficulty, highlighting its enhanced adaptability and effectiveness in handling diverse learning challenges within the dataset.

\begin{table}[ht]
    \centering
    \begin{tabular}{c|ccc}
        \toprule
        Method  & Hard  & Middle & Easy \\
        \midrule
        KD      & 48.90 & 80.53  & 92.26  \\    
        DTKD    & 51.76 & 81.88  & 93.94  \\
        \midrule
        $\Delta$ & \textcolor{ForestGreen}{+2.86} & \textcolor{ForestGreen}{+1.35} & \textcolor{ForestGreen}{+1.68} \\
        \bottomrule
    \end{tabular}
    \caption{Results on the CIFAR-100 validation with sample groups of different difficulty levels.  $\Delta$ represents the performance improvement over the classical KD. Teacher model: ResNet32$\times$4, student model: ResNet8$\times$4}
    \label{tab:hrad_mid_easy}
\end{table}

\section{Ablation}

\subsection{Reference Temperature}
For the temperature choices in the CIFAR-100 and ImageNet experiments, we strictly adhered to the temperature settings used in KD. To verify the robustness of DTKD to temperature variations, we conducted tests on CIFAR-100 using ResNet32$\times$4 as the teacher and ResNet8$\times$4 as the student, with temperatures ranging from 2 to 5. Table ~\ref{tab:temperature}   demonstrates that the performance of our method is robust against changes in the reference temperature.

\begin{table}[ht]
    \centering
    \begin{tabular}{c|cccccc}
        \toprule
        Method & T=1   & T=2   & T=3   & T=4   & T=5   & T=6  \\
        \midrule
        KD     & 73.18 & 73.75 & 73.79 & 73.50 & 73.57 & 73.29 \\    
        DTKD   & 74.0  & 75.91 & 75.97 & 76.16 & 75.46 & 75.92 \\
        \midrule
        $\Delta$ & \textcolor{ForestGreen}{+0.82} & \textcolor{ForestGreen}{+2.16} & \textcolor{ForestGreen}{+2.18} & \textcolor{ForestGreen}{+2.66} & \textcolor{ForestGreen}{+1.89} & \textcolor{ForestGreen}{+2.63}\\
        \bottomrule
    \end{tabular}
    \caption{Results on the CIFAR-100 validation with different reference temperatures. And $\Delta$ represents the performance improvement over the classical KD.}
    \label{tab:temperature}
\end{table}

\subsection{Loss Weight}
At the end of \textbf{section 3.3} of the full paper, we presented the complete loss function:

\vspace{-4mm}
\begin{align}
    \mathcal{L}_{KD} 
	=  \alpha \cdot \mathcal{L}_{\operatorname{DTKD}} + \beta  \cdot\mathcal{L}_{\operatorname{KL}}  + \gamma \cdot \mathcal{L}_{\operatorname{CE}}  \nonumber
\end{align}

We tested the influence of different configurations of $\alpha$ and $\beta$ on DTKD's performance. Table~\ref{tab:ablation} reports the student accuracy with different $\alpha$ and $\beta$. ResNet32$\times$4 and ResNet8$\times$4 are set as the teacher and the student, respectively. The $\gamma$ controls the cross-entropy (CE) weight loss, which was set to 1 in all experiments. 

\begin{table}[htp!]
    \centering
    \begin{tabular}{c|ccc}
        \toprule
        \diagbox{$\beta$}{$\alpha$} & 1     & 2     & 3              \\
        \midrule
        0.5                         & 75.13 & 75.81 & 75.87          \\
        1                           & 75.56 & 75.83 & \textbf{76.16} \\
        \bottomrule
    \end{tabular}
    \caption{Performance of DTKD on different configurations of loss weights $\alpha$ and $\beta$.}
    \label{tab:ablation}
\end{table}

\subsection{Training Process}
\begin{figure}[H]
    \centering
    \includegraphics[width=\linewidth]{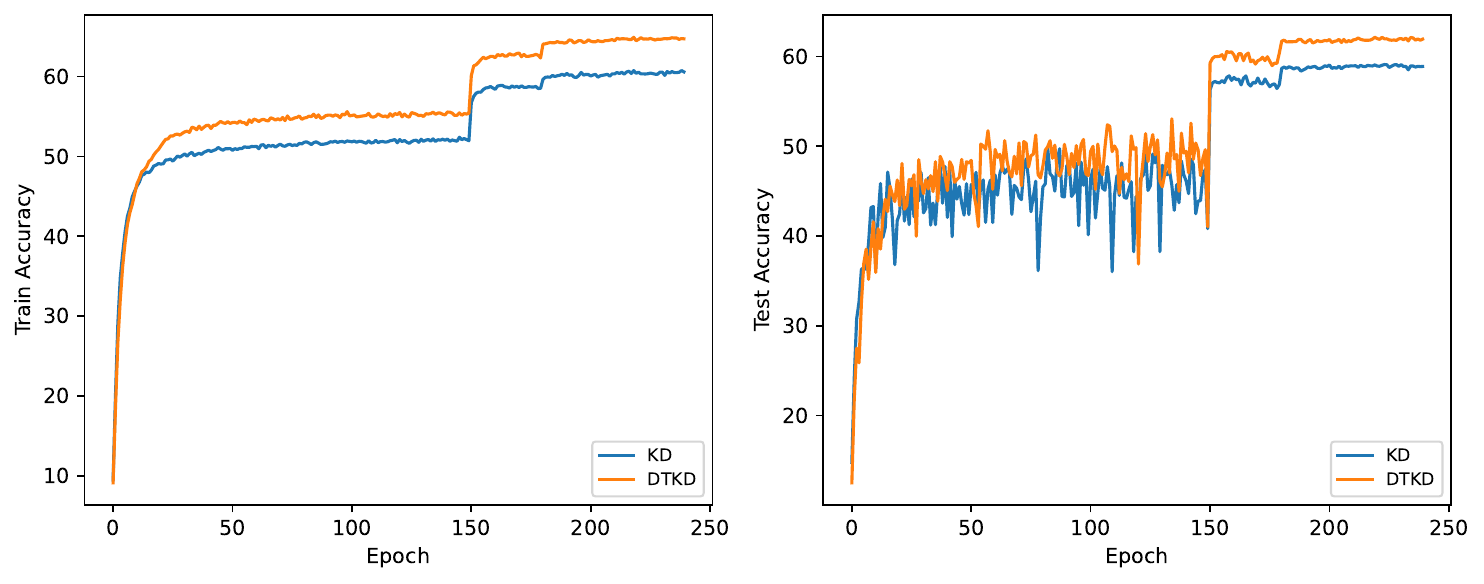}
    \caption{Accuracy of training and test sets during training of KD and DTKD. Teacher model: ResNet14 student model: ResNet8}
    \label{tab:difficult}
    \label{fig:train_test_acc}
\end{figure}

In order to clearly compare the training processes of KD and DTKD, we visualized the changes in training set accuracy and test set accuracy during the training process. Figure \ref{fig:train_test_acc} shows that our method significantly outperformed KD.

\section{Comparative analysis of DTKD and NormKD}
We note that NormKD shares a very similar perspective to ours: to eliminate the difference of sharpness for the output logits of the teacher and the student. NormKD adopts normalization on the output logits and insists on keeping a shape of normal distribution for the logits across the entire training process. Although this achieves satisfactory performance, we found that it can cause some damage to the student network itself. Besides the study in transferability shown in \textbf{section 4.4 Extensions} of the full paper where we found NormKD suffers from a cliff fall of performance, we have also made a series of further studies. In the following, we conducted distillation with various KD methods, where the teacher model is ResNet-$32\times4$ and the student model is ResNet-$8\times4$, unless stated explicitly otherwise.

\subsection{Temperature Regulation Pattern}
\begin{figure}[h]
    \centering
    \includegraphics[width=\linewidth]{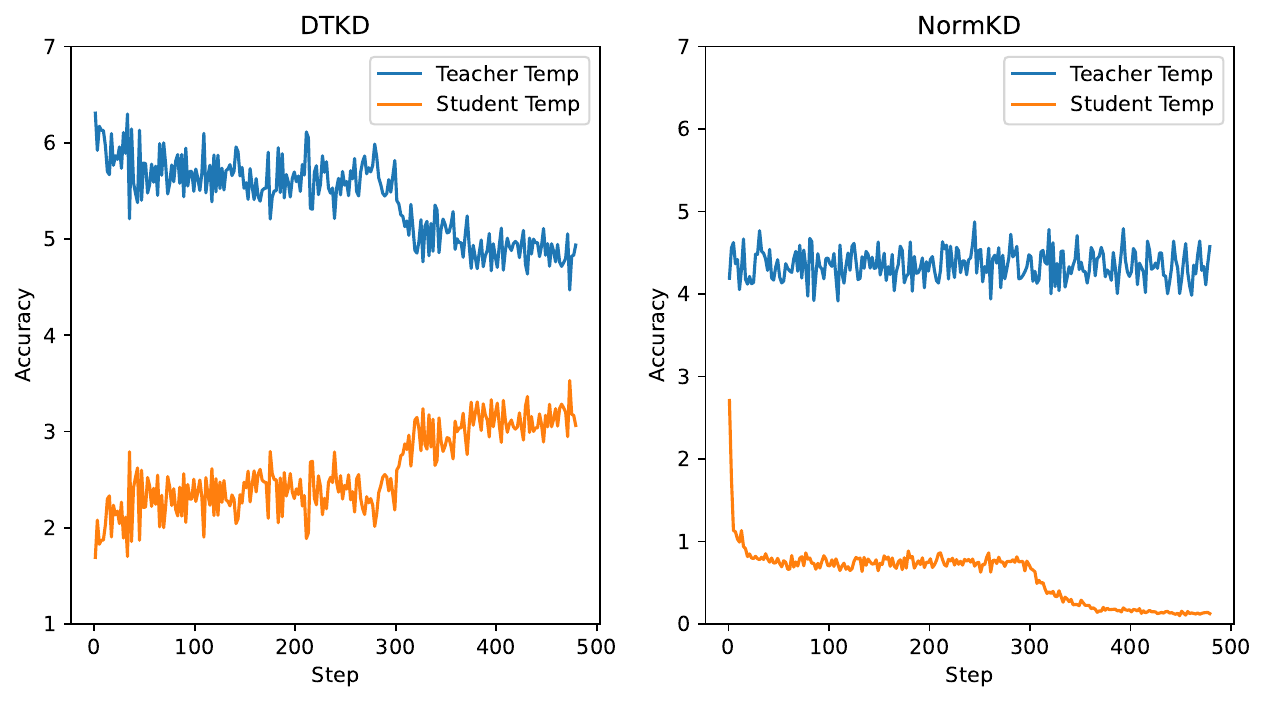}
    \caption{Temperature behaviors of DTKD and NormKD during training.}
    \label{fig:temp_normkd_dtkd}
\end{figure}

Contrasting with DTKD, NormKD employs the standard deviation of logits to determine the temperature parameter. This distinction is highlighted in our visual analysis, where we plotted the average temperature derived from the final mini-batch of each epoch. In Figure \ref{fig:temp_normkd_dtkd}, the left panel demonstrates the temperature evolution of  DTKD, while the right panel delineates the corresponding temperature changes of NormKD. This comparative visualization underscores the differential dynamics of temperature adjustment between the two methodologies. As the temperature of the student in NormKD turned to near-zero in the end, it means that the standard deviation of the student logits is also quite small. As the mean of the logits is near-zero, this indicates that the student logits also turned into near-zero values. We study this phenomenon in the next section in detail.

\subsection{Confidence analysis}
\begin{figure}[H]
    \centering
    \includegraphics[width=\linewidth]{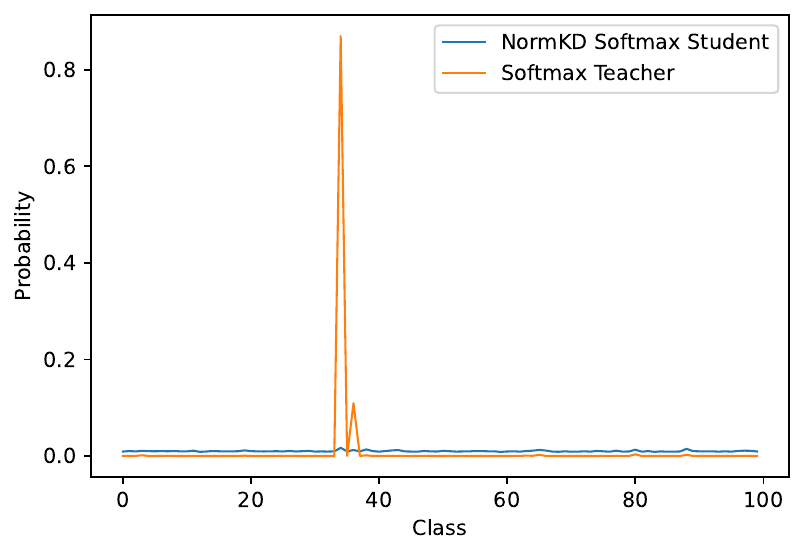}
    \caption{SoftMax prediction values of a random sample from the NormKD student model and the teacher model.}
    \label{fig:sample_softmax}
\end{figure}

We first analyzed the predictions of a random sample by a teacher and its NormKD student for comparison. As shown in Figure~\ref{fig:sample_softmax}
, we found that the NormKD student surprisingly outputs a near-uniform distribution of predictions and the maximal SoftMax prediction is still near-zero. In the following, we verified that this problem is general across the dataset.

\subsubsection{Max predictions}
\begin{figure}[H]
    \centering
    \includegraphics[width=\linewidth]{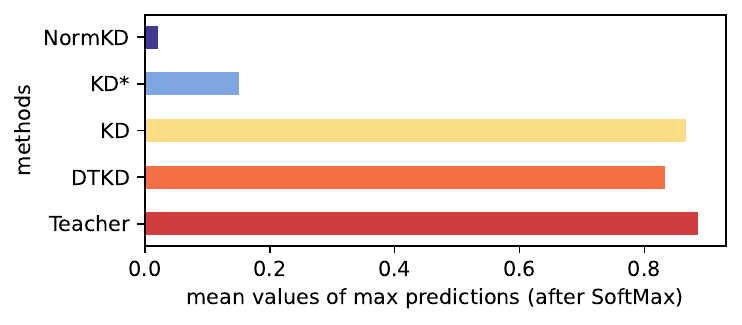}
    \caption{Averaged max predictions on 1000 sampled images of CIFAR-100 test set.}
    \label{fig:softmax}
\end{figure}

 We examined the average maximum prediction probabilities on the test set. In particular, we randomly sampled $1000$ images of the test set of CIFAR-100, and then computed the average of the maximum SoftMax predictions, as depicted in Figure~\ref{fig:softmax}. $\text{KD}^*$ denotes our adjusted version of the vanilla KD, where we altered the temperatures of the teacher and student to align more closely with those used in NormKD (i.e. teacher at 4.5 and student at 0.8). This experiment revealed that both NormKD and $\text{KD}^*$ result in average maximum prediction probabilities significantly lower than 0.2, suggesting a markedly low level of confidence in the predictions of these student models. This indicates that the logits of the NormKD student form a near-uniform distribution, resulting in the top-1 prediction being very low when the number of classes is large. Since the performance of $\text{KD}^*$ mimics that of NormKD, we suspect that it is the way of temperature regulation that caused this problem.
 
 In contrast, the KD and DTKD methods yield average maximum prediction probabilities that are more aligned with those of the teacher model, indicating a higher confidence level in the student models' predictions. This variation in prediction confidence levels is crucial as it directly impacts the reliability and effectiveness of the student models in practical scenarios. Notice that although both NormKD and our DTKD aimed at eliminating the difference in the smoothness of the models' logits distributions, the two methods walked down different paths. Whereas NormKD enforces the distributions to be Gaussian, we relaxed the shape constraint and let the temperatures be adapted dynamically. This relaxation turned out to be crucial in avoiding a near-uniform distribution.

\begin{table}[htp!]
    \centering
    \begin{tabular}{c|ccccc}
        \toprule
        Teacher & Res14 & Res32 & Res44 & Res56 & Res110 \\
        \midrule
        Conf    & 0.074 & 0.042 & 0.037 & 0.032 & 0.033 \\
        \bottomrule
    \end{tabular}
    \caption{Averaged max predictions on 1000 sampled images of CIFAR-100 test set with NormKD using different teacher models. The student model is fixed as ResNet8.}
    \label{tab:norm_res8_conf}
\end{table}

Table~\ref{tab:norm_res8_conf} shows the averaged max predictions of the ResNet8 student model distilled from various teacher models using NormKD. The issue of extremely low confidence in the student model is consistent across all cases.

\subsubsection{Principal Component Analysis}
\begin{figure}[t]
    \centering
    \includegraphics[width=\linewidth]{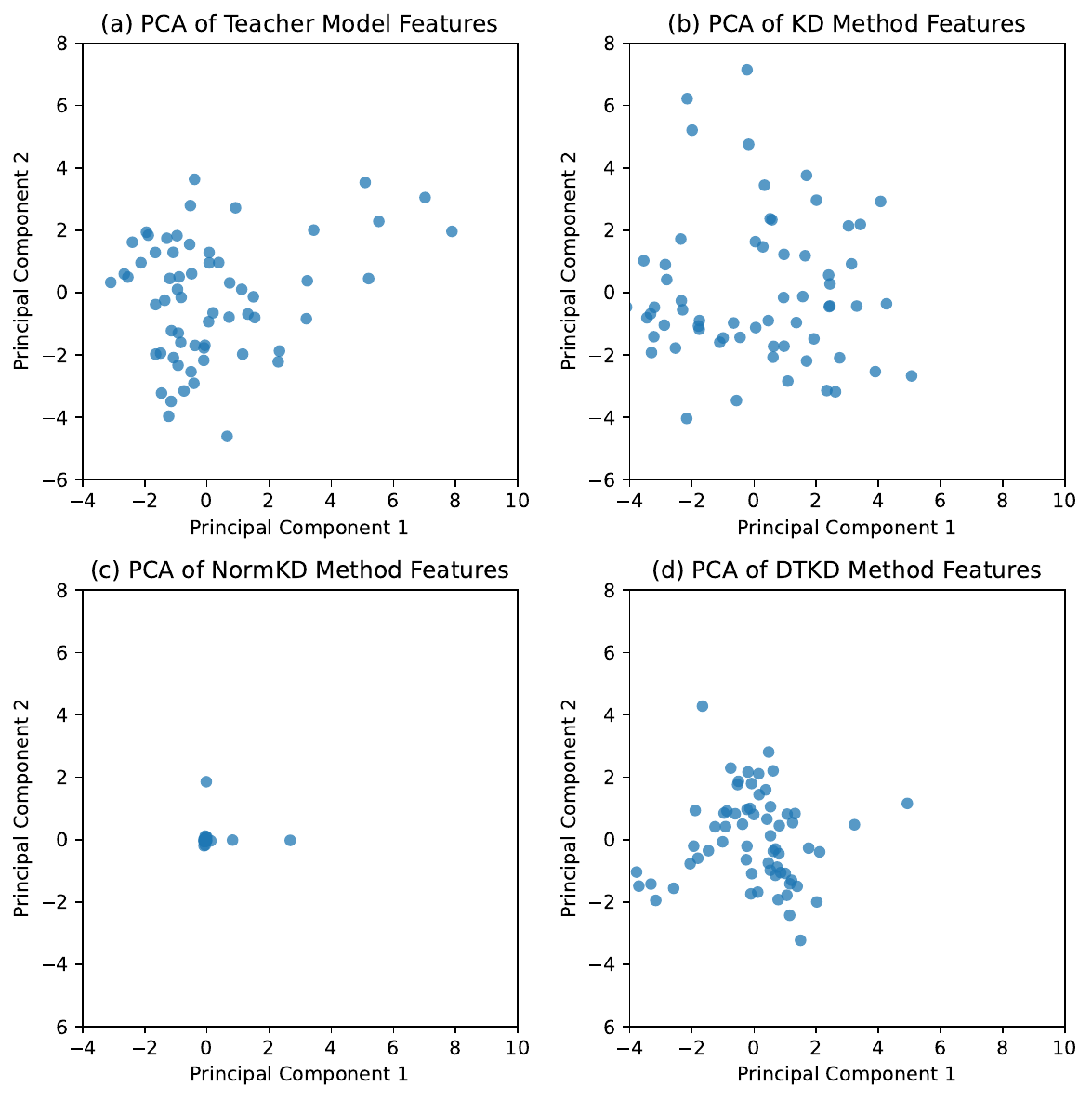}
    \caption{PCA on the features derived from: (a) teacher model,  (b) vanilla KD, (c) NormKD, and (d) DTKD respectively. }
    \label{fig:pca}
\end{figure}

A near-uniform distribution also indicates that the model is not good at discriminating similar classes, therefore leading to worse generalizability. This issue is already evident in the transferability experiments, where NormKD fell short by a big margin compared to other methods including our DTKD. Here we further conducted a Principal Component Analysis (PCA) on the features from the average pooling layer of the student models, which serves as the input for the fully connected (FC) layer. 
Figure \ref{fig:pca} shows the results, where the student model trained under NormKD shows a very tight clustering of features. This significant reduction in feature space diversity correlates with the previously noted uniform distribution of logits and suggests a diminished discriminative power, as well as the cliff fall in transferability. The features of the DTKD student model display a spread pattern that is comparable to the teacher model. The features of KD are more spread compared to both the teacher and DTKD. However, since its performance is sub-optimal, we suspect that the students learned more noises during the KD process rather than useful variance.

\subsection{Capability Gap Adaptation}
\begin{figure}[H]
    \centering
    \includegraphics[width=\linewidth]{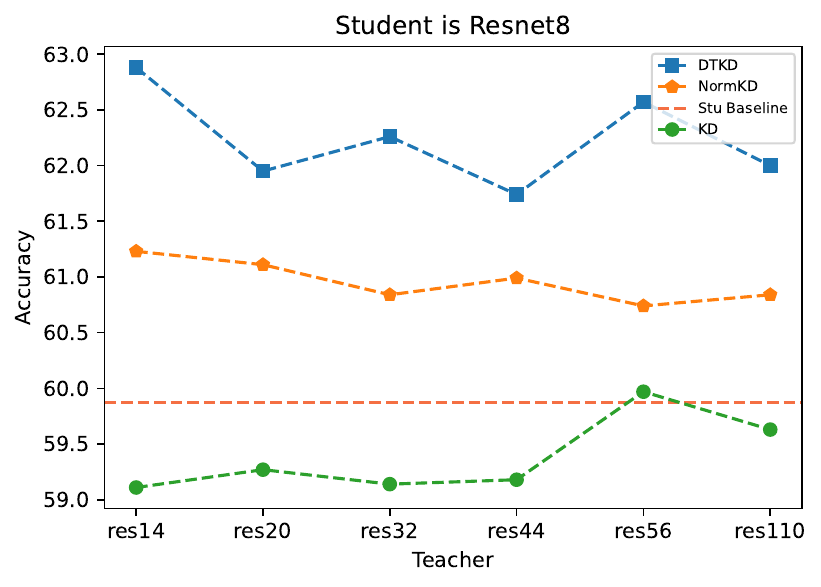}
    \caption{Performance of various KD methods when pairing the same student model with different teacher models.}
    \label{fig:norm_res}
\end{figure}

NormKD directly uses the standard deviation of teacher logits to set the teacher temperature regardless of the student outputs. As shown in Figure~\ref{fig:temp_normkd_dtkd}, this approach results in a stable teacher temperature, which does not adjust according to the student model's characteristics. Such rigidity can impede the efficiency of distillation when applied to diverse teacher-student pairings. Figure \ref{fig:norm_res} shows that with a constant student model (ResNet8) 
and various teacher models, NormKD can only achieve sub-optimal results, and the performance also decreases as the teacher model becomes larger (more capable), which is a phenomenon observed in various research works. In contrast, DTKD constantly performs better than NormKD and does not follow a trend of performance degradation as the teacher becomes larger. 

\end{document}